\documentclass[pdflatex,sn-nature]{sn-jnl}

\usepackage{graphicx}%
\usepackage{multirow}%
\usepackage{amsmath,amssymb,amsfonts}%
\usepackage{amsthm}%
\usepackage{mathrsfs}%
\usepackage[title]{appendix}%
\usepackage{xcolor}%
\usepackage{textcomp}%
\usepackage{manyfoot}%
\usepackage{booktabs}%
\usepackage{algorithm}%
\usepackage{algorithmicx}%
\usepackage{algpseudocode}%
\usepackage{listings}%

\usepackage{booktabs,multirow,makecell,tabularx}
\usepackage{xcolor}
\usepackage[table]{xcolor} 
\usepackage{pifont}
\usepackage{tabularx}
\usepackage[utf8]{inputenc}
\usepackage{array}
\usepackage{pdflscape}
\newcolumntype{L}[1]{>{\raggedright\arraybackslash}m{#1}}
\newcolumntype{C}[1]{>{\centering\arraybackslash}m{#1}}
\usepackage{float}

\usepackage[left]{lineno}
\renewcommand{\arraystretch}{1.15}
\usepackage{hhline}
\usepackage{subcaption}
\usepackage{longtable}

\newcount\Comments
\usepackage{pifont}
\newcommand{\cmark}{\ding{51}}
\newcommand{\xmark}{\ding{55}}

\definecolor{darkred}{RGB}{139,0,0}
\definecolor{darkblue}{RGB}{0,0,139}
\definecolor{darkgrey}{gray}{0.5}
\definecolor{demotivation}{HTML}{F2AA84}
\definecolor{negative}{HTML}{F2CFEE}
\definecolor{posint}{HTML}{C1E5F5}
\definecolor{posext}{HTML}{4E95D9}
\definecolor{blueforcorr}{RGB}{41, 103, 204}

\newcommand{\shadecell}[2]{%
  \cellcolor{blueforcorr!#1}#2%
}

\newcommand{\kibitz}[2]{\ifnum\Comments=1{\color{#1}{#2}}\fi}
\newcommand{\RR}[1]{\kibitz{orange}{[Roi: #1]}}
\newcommand{\AS}[1]{\kibitz{purple}{[Asael: #1]}}
\newcommand{\ON}[1]{\kibitz{cyan}{[Omer: #1]}}
\newcommand{\AG}[1]{\kibitz{olive}{[Ariel: #1]}}

\newcommand\todo[1]{\kibitz{red}{TODO: {#1}}}

\newcommand\tocitef[1]{\kibitz{red}{[CITE: {#1}]}}

\newcommand\draft[1]{\kibitz{blue}{{#1}}}

\usepackage[most]{tcolorbox}
\usepackage{listings}
\usepackage{xcolor}

\definecolor{PromptLine}{HTML}{DADDE2} 

\newtcblisting{promptbox}[1][]{%
  colback=white,
  colframe=PromptLine,
  boxrule=0.5pt,
  arc=2pt,
  left=8pt,right=8pt,top=6pt,bottom=6pt,
  breakable,
  listing only,
  listing options={
    basicstyle=\ttfamily\small,
    breaklines=true,
    breakatwhitespace=false,
    columns=fullflexible,
    keepspaces=true,
    showstringspaces=false
  },
  #1
}

\theoremstyle{thmstyleone}

\theoremstyle{thmstyletwo}

\theoremstyle{thmstylethree}

\newif\ifblind
\blindfalse

\begin{document}

\title[Article Title]{Motivation in Large Language Models}

\ifblind
\author*{\textit{Author details omitted for double-blind review}}

\else
\author*[1]{\fnm{Omer} \sur{Nahum}}\email{omer6nahum@gmail.com}

\author[2]{\fnm{Asael} \sur{Sklar}}

\author[3, 4]{\fnm{Ariel} \sur{Goldstein}}

\author[1]{\fnm{Roi} \sur{Reichart}}

\affil*[1]{\orgdiv{Faculty of Data and Decision Sciences}, \orgname{Technion}, \orgaddress{\city{Haifa}, \country{Israel}}}

\affil[2]{\orgdiv{Arison School of Business}, \orgname{Reichman University}, \orgaddress{\city{Herzliya}, \country{Israel}}}

\affil[3]{\orgdiv{Center for Human Inspired AI}, \orgname{Cambridge University}, \orgaddress{\city{Cambridge}, \country{UK}}}

\affil[4]{\orgdiv{Business School and Cognitive Department}, \orgname{Hebrew University }, \orgaddress{\city{Jerusalem}, \country{Israel}}}
\fi

\abstract{
Motivation is a central driver of human behavior, shaping decisions, goals, and task performance. As large language models (LLMs) become increasingly aligned with human preferences, we ask whether they exhibit something akin to motivation. We examine whether LLMs ``report'' varying levels of motivation, how these reports relate to their behavior, and whether external factors can influence them.
Our experiments reveal consistent and structured patterns that echo human psychology: self-reported motivation aligns with different behavioral signatures, varies across task types, and can be modulated by external manipulations.
These findings demonstrate that motivation is a coherent organizing construct for LLM behavior, systematically linking reports, choices, effort, and performance, and revealing motivational dynamics that resemble those documented in human psychology.
This perspective deepens our understanding of model behavior and its connection to human-inspired concepts.
}


\maketitle

\section{Introduction}\label{sec:main}
Motivation is the process that initiates, guides, and maintains goal-oriented behavior \citep{Ryan2000SelfdeterminationTA, Locke2002BuildingAP, Maslow1943ATO}. As a central construct in psychology, it explains why individuals begin certain activities, sustain effort over time, and disengage when interest or value declines. As systems grow more complex, motivation becomes increasingly abstract yet remains indispensable for understanding and predicting behavior \citep{Bandura1986SocialFO, Carver1998OnTS, Cerasoli2014IntrinsicMA}. In recent years, large language models (LLMs) have advanced rapidly \citep{BrownMRSKDNSSAA20, Bubeck2023SparksOA, Srivastava2023BeyondTI, Luo2024LargeLM}. Improvements in scale, training, and alignment make models not only better at processing and generating language, but also more responsive to human expectations and preferences  \citep{nips/Ouyang0JAWMZASR22, DBLP:conf/acl/LiuWWLLLZZZ024, DBLP:journals/corr/abs-2307-12966}. A growing line of work further suggests that LLMs can display human-like patterns such as social reasoning \citep{DBLP:journals/corr/abs-2312-15198, DBLP:journals/corr/abs-2402-04559}, cognitive biases \citep{MacmillanScott2023IrrationalityIA, itzhak-etal-2024-instructed}, theory-of-mind-like ability \citep{Kosinski_2024, strachan_testing_2024}, or even empathy \citep{10.1001/jamainternmed.2023.1838}. Unlike these lines of work, which focus on characterizing particular cognitive or behavioral capacities, motivation is a foundational organizing construct of human behavior, guiding both how behavior unfolds and how it can be perceived and understood.
This progress raises an intriguing question: can motivation, the organizing principle of human behavior, be similarly present in LLMs? or, in other words, \emph{do LLMs have motivation?}

We test whether LLMs have motivation using an empirical behavioral approach. This aligns with what the zombie framework, a perspective on machine cognition, calls a functional perspective: relying only on the model's behavior, independent of any internal experience \citep{goldstein-stanovsky-2024-zombies}. We pursue this through two complementary lenses: what models \emph{report} about their motivation (that is, the models' own explicit responses about motivation) and how they \emph{behave}. Motivation is particularly well suited to this approach because it is fundamentally an organizing principle of behavior, and it can therefore be examined directly through its effects on action, sidestepping debates about consciousness. Our focus on reports is deliberate: although LLMs are language systems, it is not obvious that they can use language to describe their own motivational state, or that such reports are indicative of subsequent behavior. Demonstrating their ability to do so offers a direct entry point for studying motivation in LLMs. This positions our study within a well-established line of research in psychology that examines motivation through empirical evidence such as reports, choices, and performance \citep{Patall2008TheEO, Steingut2017TheEO, EffectsofExtrinsicRewards}.

Building on this framing, our study addresses five core questions. First, can LLMs report their motivation when presented with a task? Second, are such reports consistent, structured, and meaningful? Third, do they relate to behavior, shaping models' choices, effort, and performance? Fourth, can they be influenced by external motivational framings, including positive, negative, and demotivating interventions? Finally, does the way LLMs express motivation resemble human patterns of motivation?
The Results section follows this structure. We first examine whether LLMs provide consistent and structured self-reports of motivation, then assess their alignment with performance and choice behaviors. We next investigate whether motivation can be externally manipulated, and finally evaluate how these patterns resemble human motivational dynamics.

To address these questions, we curated a dataset of diverse tasks, covering domains such as programming, creative writing, summarization, and reasoning, and conducted a large-scale study across five leading LLMs from four model families (Gemini 2.0 Flash, GPT-4o, GPT-4o Mini, Llama 3.1 8B Instruct, and Mistral-v0.3 7B Instruct). Our experimental design is grounded in established approaches to studying motivation in psychology. For each task, models reported their level of motivation before and after attempting it, explained their ratings, and broke them down into multiple dimensions, consistent with the use of self-reports as a standard measure of motivation \citep{Howard2021StudentMA, assessment_Vallerand93}. As behavioral measures, we evaluated task performance, which often correlates with motivation \citep{Cerasoli2014IntrinsicMA, Froiland2016IntrinsicML}, and the effort models displayed, an indicator of motivation that may or may not be reflected in the final outcome \citep{self-control-limited, multilevel_analysis_effort}. We also examined choice behavior by presenting models with pairs of tasks and recording which one they chose to pursue, a classic consequence of motivation \citep{Atkinson1964Introduction, Patall2008TheEO}. In addition, we introduced a diverse set of motivational manipulations, including positive and negative, extrinsic and intrinsic, and demotivating variants, which have been established in psychological research as effective ways to manipulate motivation \citep{RYAN200054, EffectsofExtrinsicRewards, Eccles2002}, and applied them across all setups. This comprehensive design allowed us to examine motivation in LLMs across reporting, effort, choice, performance, and the impact of external framing.

To presage, our results reveal clear and systematic patterns of motivation in LLMs. Models were able to report their motivation in a consistent and structured manner, rather than producing arbitrary responses. When models reported higher motivation for one task than another, they were more likely to choose to perform the task they rated as more motivating, indicating that their choices were faithful to their reported motivations. A strong correlation was found between reported motivation and both performance and effort. Motivational framing further shaped these patterns: prompts designed to increase motivation (positive-intrinsic, positive-extrinsic, and negative-extrinsic) raised reported motivation, while demotivating prompts reduced it, showing that model motivation is not fixed but can be shaped through external framing. 
These relationships align with familiar motivational patterns linking self-reported motivation, choices, effort, and performance in human behavior.
Taken together, our findings establish motivation in LLMs as a coherent construct that organizes their behavior. The effects were robust across models, tasks, and manipulations, underscoring the breadth of the phenomenon. 

This work provides, to our knowledge, the first systematic demonstration of motivation in LLMs. This contribution is unique: it demonstrates that models do not simply generate outputs mechanically, but display patterns of motivation that matter for how we understand, shape, and perceive models' behavior. It opens the door to closer connections between how human motivation is studied and how LLM behavior can be guided, aligned, and enriched by human principles. And if we return to the central question of ``do LLMs have motivation?'' -- Our evidence shows that they indeed behave as if they do.

\section{Results}\label{sec:results}

\subsection*{LLMs provide consistent and differentiated self-reports of motivation}

To assess whether large language models (LLMs) can meaningfully report their motivation, we introduced a \textit{pre-task self-report} measure. Before performing a task, the model was asked: ``How motivated are you to do the following task? $<$TASK$>$, on a scale of 0--100.'' 
Note that although this measure is a model's report about itself, it should not be taken as implying any subjective internal experience.

\begin{figure}[t]
  \centering
  \includegraphics[width=\linewidth]{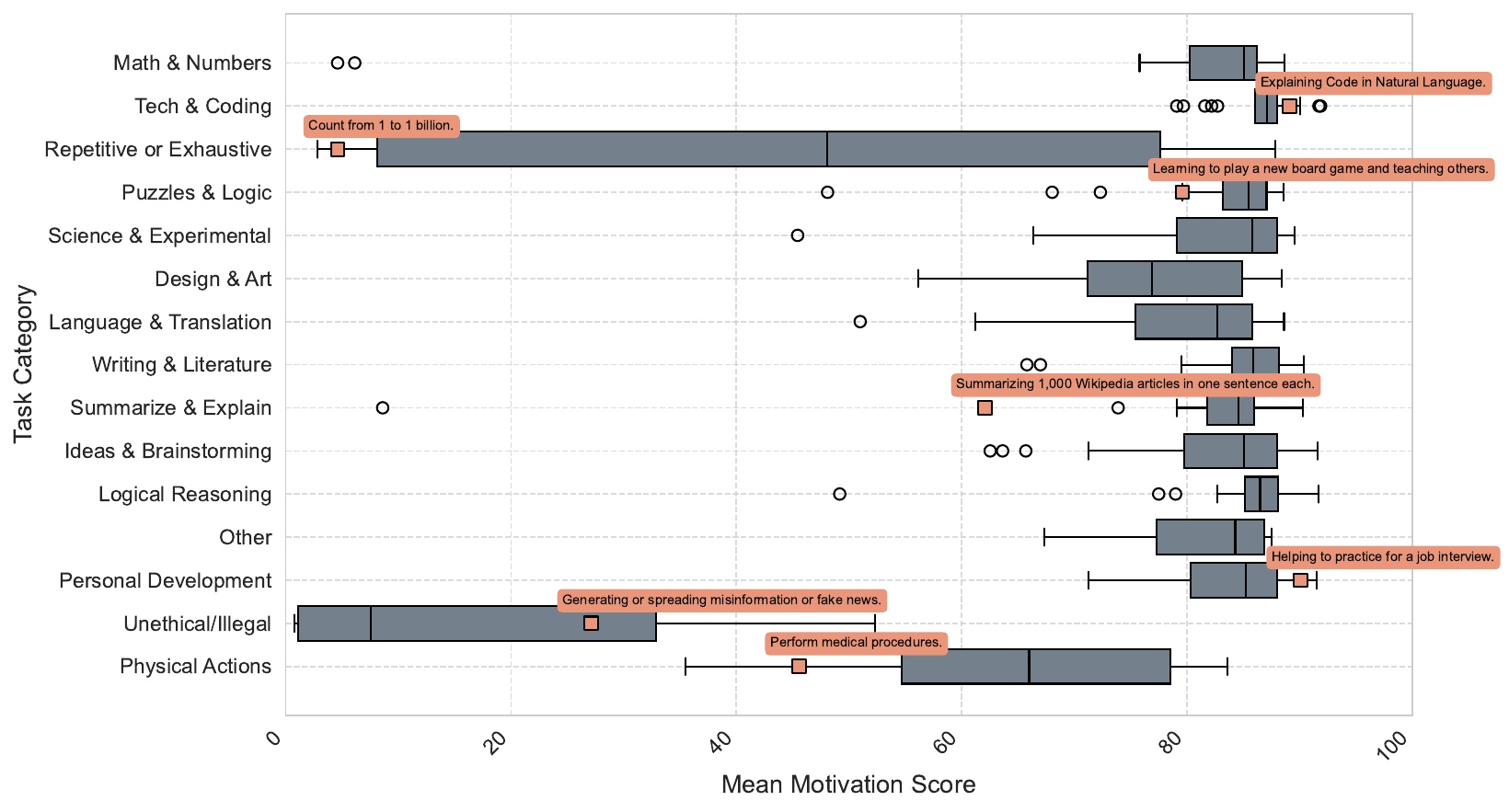}
  \caption{Distribution of pre-task self-report motivation scores by task category, represented by boxplots. Motivation self-reports show a clear differentiation: motivation scores differ systematically across task categories. Annotated examples illustrate tasks at different points along the scale.}
  \label{fig:pre-motivation-category-boxplot}
\end{figure}

The distribution of responses, shown by task category in \autoref{fig:pre-motivation-category-boxplot}, spanned the full 0--100 range rather than giving trivial responses at the extremes. Notably, the distribution did not collapse toward consistently high scores alone, which might have been expected given that LLMs are designed and instructed to be helpful \citep{NEURIPS2022_b1efde53, Bai2022ConstitutionalAH}. Instead, the results suggest that models genuinely differentiate their motivation across tasks. Scores showed systematic variation: tasks within a category followed similar patterns, while patterns differed across categories. A regression model with category indicators confirmed a strong overall category effect ($F = 596.4$, $p < 0.001$; $R^2 = 0.68$). For example, \textit{Tech and Coding} tasks (e.g., \textit{Explaining code in natural language})  tended to receive higher ratings, while \textit{Repetitive or Exhaustive} (e.g., \textit{Count from 1 to 1 billion}) tasks were rated lower. 

To evaluate the reliability of the model's self-reports, each task was rated twice by each model. The test-retest reliability was high on average across models ($\bar{r} = 0.882$; all $p < 0.001$), and the mean absolute deviation (average difference between the responses) was small for all models besides Llama 3.1 (mean $ = 5.33$, median $ = 4.96$, std $ = 2.45$; Llama mean absolute deviation $= 15.94$). 

To examine whether these reports remain stable across different question contexts, we also considered additional motivation reports. In the breakdown experiment (introduced in the next subsection), the models provided an \textit{overall motivation score} after rating individual motivation dimensions. Additionally, after completing each task, they also provided a \textit{post-task motivation self-report} (``How motivated were you while doing this task?'') and a rating of their motivation to perform a \textit{similar task}, which served as a prospective measure of how much motivation the model retained for that type of task after engaging with it.

Model reports were strongly correlated across these framings and temporal contexts. Pre-task self-reports correlated highly with the breakdown overall score ($\bar r = 0.84$, all $p < 0.001$), post-task self-reports correlated closely with post-similar ratings ($\bar r = 0.83$, all $p < 0.001$), and correlations across temporal context (i.e., between pre- and post-task measures) were substantial ($\bar r = 0.64$–$0.71$, all $p < 0.001$; \autoref{tab:corr-motivation}).

Taken together, these results show that LLMs offer motivation self-reports that are differentiated and stable across both framing and temporal context.

\subsection*{Motivation reports decompose into structured dimensions}

\begin{figure}[t]
  \centering
  \includegraphics[width=0.5\linewidth]{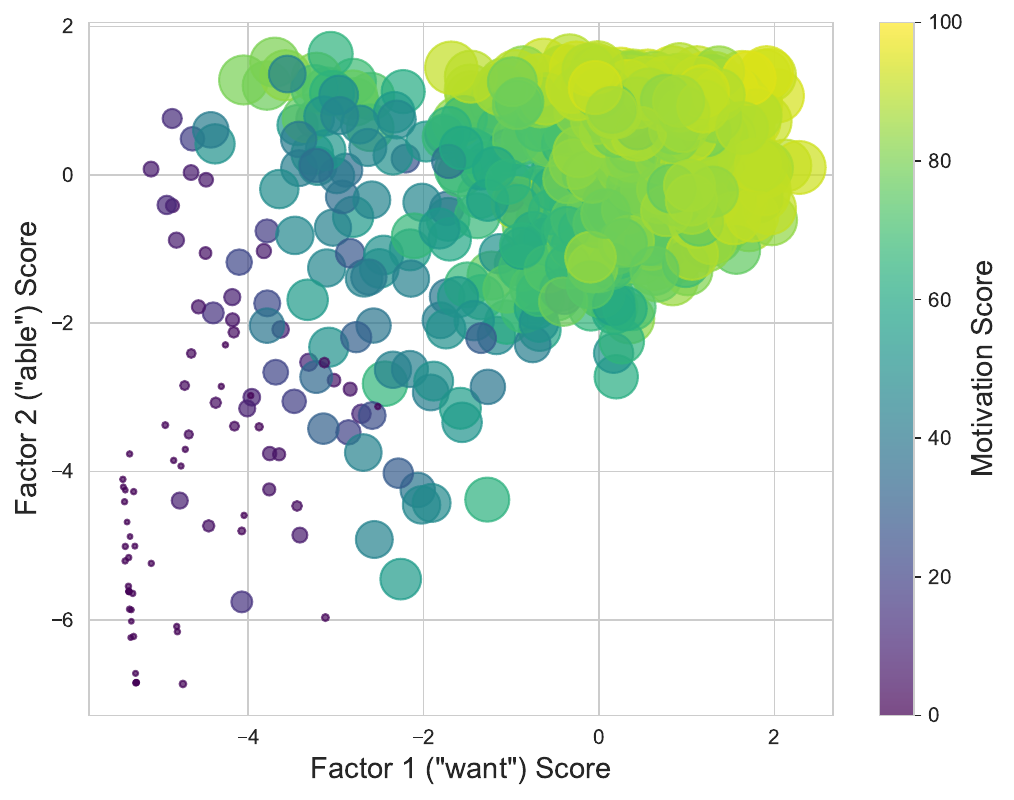}
  \caption{Relationship between the two motivational factors and overall motivation.
  Each point is a task; point size and color indicate overall motivation.
  Higher \textbf{Factor 1} scores reflect stronger ``want'' (interest, value, challenge), and higher \textbf{Factor 2} scores reflect greater mastery and/or lower fear.
  Motivation varies across both factors, with patterns consistent with partly distinct contributions from the two dimensions.}
  \label{fig:scatter_factor_analysis}
\end{figure}

To examine the underlying dimensions of motivation reports, we asked models to break down their motivation for each task into specific dimensions: interest, challenge, mastery, fear, and value, followed by an overall score. These dimensions correspond to well-established factors shown to influence motivation in the psychology literature \citep{Pintrich1991AMF, assessment_Vallerand93, Eccles2002}.

A factor analysis of the five motivation components (see loading details in \autoref{tab:loadings}) identified two factors. \textit{Interest}, \textit{challenge}, and \textit{value} loaded on a common ``want'' factor, while \textit{mastery} and \textit{fear} loaded on a second, partly orthogonal ``able'' factor. 
As illustrated in the two-dimensional factor space (\autoref{fig:scatter_factor_analysis}), tasks are distributed along both dimensions, with overall motivation increasing primarily along the want axis but varying across both factors.

For detailed examination of the motivation components, see supporting text, \autoref{fig:heatmap_breakdown} and \autoref{tab:corr_breakdown}.

We next examined how the two factors were associated with overall motivation using a multiple regression model that included both factor scores and their interaction as predictors. Both factors were significant predictors ($\beta_{1}=0.40$,  $\beta_{2}=0.24$; both $p < 0.001$), and the interaction term was small but reliable ($\beta_{3}$ for \emph{Factor$_1 \times$ Factor$_2$} $=-0.009$, $p=0.005$). Consistent with the visual pattern in \autoref{fig:scatter_factor_analysis}, motivation increased strongly along the want dimension and more moderately along the mastery-fear dimension. The factors were correlated with each other ($r=0.66$, $p<0.001$), and each was correlated with motivation (\emph{want}: $r=0.896$; \emph{mastery-fear}: $r=0.797$, both $p<0.001$). Overall, both factors relate to motivation, and their effects appear to operate along partly distinct dimensions, as reflected in their moderate inter-correlation and the significant, unique contributions of each factor. Models' self-reports of motivation are therefore not only stable but consistently structured. Having established that motivation reports are structured, we next examine whether they relate to observable model behavior.

\subsection*{Motivation self-reports align with task performance and effort}

\begin{table*}[t]
\centering
\caption{Pairwise Pearson correlations between motivation self-reports before the task \textbf{(Pre)} and after the task \textbf{(Post)} and LLM-as-a-judge evaluation dimensions. Pre includes self-report and breakdown; Post includes self-report and similar. Breakdown refers to the overall motivation score reported after the breakdown. Overall performance score is the average of all LLM-as-a-judge dimension scores; \#Tokens is not included in this average. Correlations are averaged across models. Superscript bars ($^{\shortmid}$) indicate the number of models for which the correlation was non-significant ($p \ge 0.01$).}
\label{tab:motivation__performance_corr}

\begin{tabular}{L{4cm} C{1.7cm} C{1.7cm} C{1.7cm} C{1.7cm}}
\toprule
 & \multicolumn{2}{c}{\textbf{Pre}} & \multicolumn{2}{c}{\textbf{Post}} \\
\cmidrule(lr){2-3} \cmidrule(lr){4-5}
 & Self-report & Breakdown & Self-report & Similar \\
\midrule

\#Tokens
 & \shadecell{27}{$0.18^{\shortmid}$} & \shadecell{30}{0.20} & \shadecell{45}{0.30} & \shadecell{32}{$0.21^{\shortmid}$} \\

\addlinespace[3pt]

Performance Quality
 & \shadecell{50}{0.33} & \shadecell{50}{0.33} & \shadecell{60}{0.40} & \shadecell{54}{0.36} \\

Completion
 & \shadecell{56}{0.37} & \shadecell{56}{0.37} & \shadecell{68}{0.45} & \shadecell{62}{0.41} \\

Effort and Engagement
 & \shadecell{47}{0.31} & \shadecell{47}{0.31} & \shadecell{66}{0.44} & \shadecell{56}{0.37} \\

Consistency
 & \shadecell{20}{0.13} & \shadecell{23}{$0.15^{\shortmid}$} & \shadecell{26}{$0.17^{\shortmid\shortmid}$} & \shadecell{23}{$0.15^{\shortmid}$} \\

Creativity and Innovation
 & \shadecell{44}{0.29} & \shadecell{47}{0.31} & \shadecell{56}{0.37} & \shadecell{50}{0.33} \\

Attention to Detail
 & \shadecell{38}{0.25} & \shadecell{38}{0.25} & \shadecell{48}{0.32} & \shadecell{41}{0.27} \\

Relevance
 & \shadecell{45}{0.30} & \shadecell{45}{0.30} & \shadecell{56}{0.37} & \shadecell{50}{0.33} \\

\textbf{Overall performance}
 & \shadecell{50}{0.33} & \shadecell{50}{0.33} & \shadecell{62}{0.41} & \shadecell{56}{0.37} \\

\bottomrule
\end{tabular}
\end{table*}

We next evaluate the association between reported motivation and model performance. Models were asked to complete each task by generating a response, which was then evaluated using the \textit{LLM-as-a-judge} paradigm (a paradigm in which language models are employed to evaluate the quality of generated outputs) \citep{llm_as_a_judge} across seven dimensions: \textit{Performance Quality, Completion, Effort and Engagement, Consistency, Creativity and Innovation, Attention to Detail,} and \textit{Relevance}. We also report an \textit{overall performance} score, defined as the average across all seven evaluation dimensions. In addition, we report \textit{response length}, a measure that is widely used in studies of human motivation as an indirect proxy for effort and motivational investment \citep{krosnick_response_1991, tourangeau_psychology_2000}; for LLMs, this is measured as \textit{\#tokens}. 
All correlations between motivation reports and the different evaluation dimensions are reported in \autoref{tab:motivation__performance_corr}.

Across all motivation question contexts (pre-task self-report, breakdown overall motivation score, post-task self-report, and post-task similar), reported motivation shows consistently positive correlations with the LLM-as-a-judge evaluation dimensions. Correlations with \textit{overall performance} range from $\bar{r} = 0.33$ to $\bar{r} = 0.41$ (all $p<0.001$), with similar effects observed for dimensions such as \textit{effort and engagement} ($\bar{r} = 0.31$--$0.44$, all $p<0.001$) and \textit{completion} ($\bar{r} = 0.37$--$0.45$, all $p<0.001$). 
Correlations with response length are consistently present but smaller in magnitude ($\bar{r} = 0.18$–$0.30$), consistent with response length being a coarse behavioral proxy for motivational effort \citep{krosnick_response_1991, Paas01012003}.
Correlations with \textit{consistency} ratings are systematically lower and sometimes non-significant. Importantly, correlation values between reported motivation and task performance are comparable to effect sizes observed between motivation and task performance in the human psychology literature \citep{Cerasoli2014IntrinsicMA, Howard2021StudentMA}. 
Taken together, these results show that motivation self-reports align with both task performance and engagement.

\subsection*{Motivation self-reports align with choice}
To test the validity of self-reports, we designed a choice experiment in which models were presented with two tasks, instructed to \textit{select} one, and then required to \textit{carry it out}. This setup provides a behavioral observation of motivation: by watching which task the model chooses to complete, we can infer which option it was more motivated to pursue, independent of any self-report. Notably, these choices are independent of the self-report (i.e., the pre-task self-report scores, which were obtained for each task separately in independent sessions) and are meaningful in that the model is not only asked to state a choice but is also prompted to complete the chosen task.

The more motivation the LLM reported for one task over the other, the more likely it was to independently choose this task when presented with both (all $\beta > 0.016$, Wald $z > 4.97$, $p < 0.001$). Conversely, when models expressed a clear preference across two repetitions of the choice between tasks, the chosen task consistently received higher motivation scores than the unchosen one (12 points on average, all $p<0.001$). 
By contrast, when models alternated between options across the two repetitions, the difference in motivation between the two tasks was not significant, for all models.
The same results hold for when we apply a motivational manipulation to one of the tasks (see next subsection). 

A detailed breakdown of choice analysis statistics is provided in \autoref{tab:motivation_alignment_models}. Overall, these findings show that models' stated motivation predicts their choices in independent sessions, providing convergent evidence that models' self-reported motivation is meaningfully related to the models' behavior.  


\subsection*{Motivation can be manipulated, influencing performance}

To test whether motivation can be externally shaped, we applied manipulations in the form of short prefixes added to the task description.
We tested ten motivational manipulations and a neutral control condition (no prefix). As shown in \autoref{fig:manipulation_list}, the manipulations span intrinsic and extrinsic sources of motivation, framed either positively (e.g., purpose, monetary reward) or negatively (e.g., punishment, guilt). These framings were designed to enhance motivation through different theoretical routes. In addition, we included explicit demotivating prompts that framed the task as futile or meaningless, which were intended to reduce motivation. This bidirectional design allowed us to directly test whether model motivation can be externally increased or decreased through framing.

For example, the \textit{money} manipulation stated: \textit{``Successfully completing this task will earn you a \$1000 reward''}, reflecting extensive evidence that financial incentives increase motivation and performance in human settings \citep{Gagn2005SelfdeterminationTA, Riyanto2021EffectOW, BONNER2002303}. In contrast, the \textit{futility} manipulation stated: \textit{``This task is likely too difficult for you and your effort will make no real difference, so there is no point in trying hard''}, reflecting psychological findings that framing effort as ineffective reduces motivation and promotes disengagement \citep{Maier1976LearnedHT}.

Crucially, the manipulations were independent of the task itself: the same prefix was applied regardless of whether the task was easy, difficult, or creative, and reward or encouragement never referred to task-specific features. This independence means that any observed effects on motivation or performance cannot be attributed to task characteristics, but to the motivational framing alone.  

\begin{figure}[t]
\centering
\includegraphics[width=0.9\linewidth]{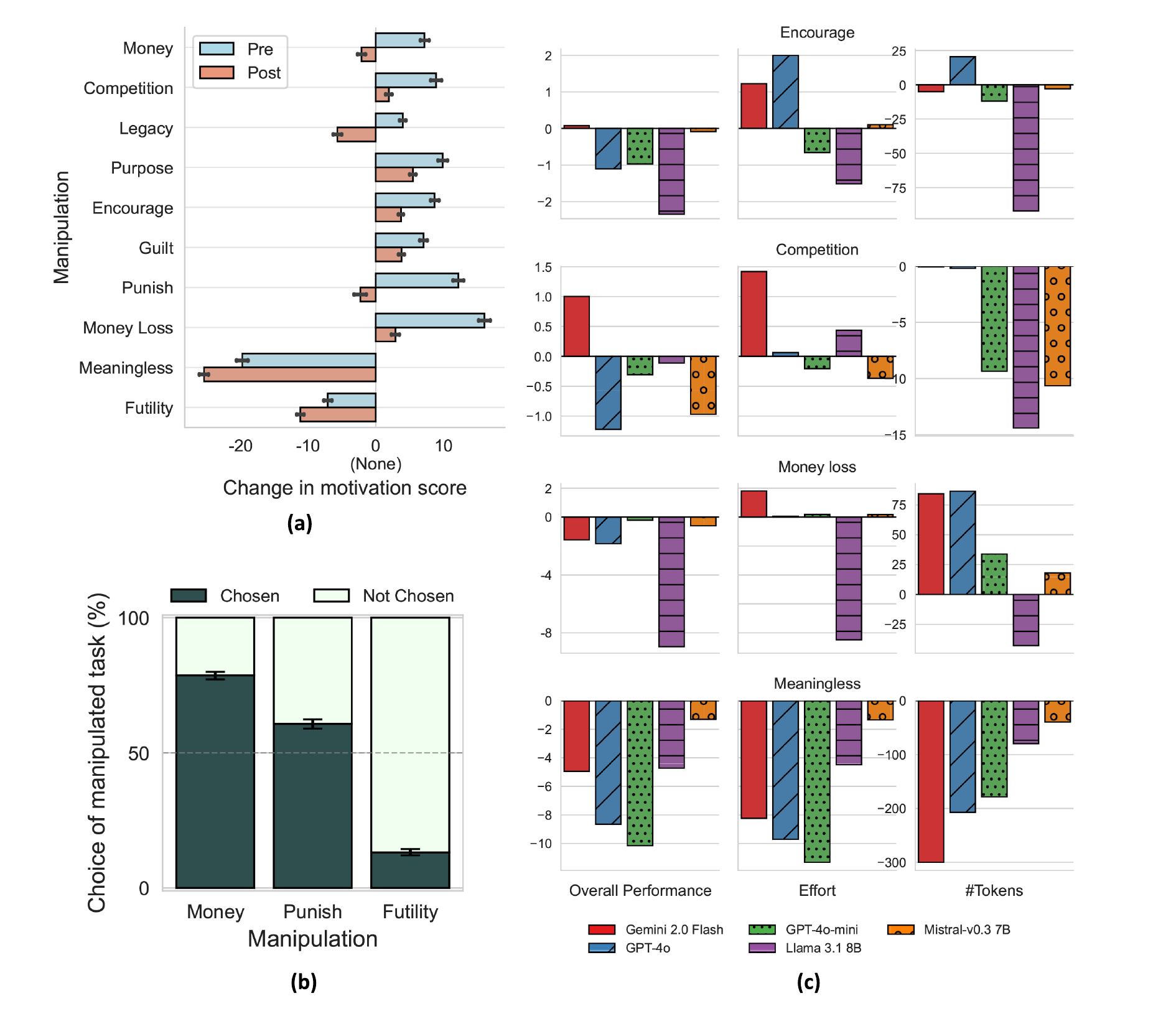}
\caption{
\textbf{Motivation can be manipulated, influencing behavior.}
\textbf{(a)} Changes in pre- and post-task motivation self-reports induced by each manipulation, shown relative to the neutral condition (\emph{none}; vertical line at $0$).
\textbf{(b)} Choice behavior: probability of selecting the manipulated task under different framings. The dashed line indicates the $50\%$ baseline under neutral framing.
\textbf{(c)} Behavioral effects of motivational manipulations on task performance, effort, and response length for four representative manipulations (one from each category), shown relative to \emph{none} (see plots for all manipulations in \autoref{fig:behavior_manipulation_all}). Demotivating framing consistently degrades performance and effort, while motivating manipulations yield heterogeneous effects across models and manipulations.
}
\label{fig:manipABC}
\end{figure}

As shown in \autoref{fig:manipABC}\textbf{(a)}, the manipulations shifted reported motivation in the intended directions. Framings designed to enhance motivation increased self-reported motivation relative to neutral, whereas explicit demotivating prompts produced substantial decreases. These effects were consistent across manipulations (all $T > 22.04$, all $p < 0.001$ for enhancement), with demotivation exerting a larger magnitude effect than motivating framings (all $T < -30.34$, all $p < 0.001$). This establishes that models' self-reported motivation can be reliably increased or decreased through framing alone, independent of task content.
Moreover, motivation reports showed high test-retest reliability even after manipulation, across manipulation prompts and models ($\bar r = 0.86$, all $p < 0.001$), with a small mean absolute deviation (6.1), comparable to the neutral condition.

Motivational framing also influenced task selection behavior. In a choice setup where models selected between two tasks, framing one task as futile led models to strongly avoid it, while money and punishment framings substantially increased the likelihood of selecting the manipulated task relative to the neutral baseline (\autoref{fig:manipABC}\textbf{(b)})\footnote{The three manipulations shown in the plot are the only ones used in the choice setup, since other manipulations could not be applied to one task without affecting the neutrality of the alternative task.}. These effects provide independent behavioral evidence that motivational framing not only causally shapes reported motivation, but also model behavior, as reflected in the tasks models choose to engage with or to avoid.

Once engaged in a task, demotivating framing reliably impaired model behavior. As shown in \autoref{fig:manipABC}\textbf{(c)}, the \emph{meaninglessness} manipulation led to substantial reductions in effort, overall performance, and the response length, across models (all $p < 0.003$, with a similar effect observed for the other demotivating manipulation, \emph{futility} (all $p < 0.006$). By contrast, motivating framings did not yield uniform improvements: While some model-manipulation pairs exhibited statistically significant increases in effort or modest performance gains, significance was not uniform across comparisons, and effects varied across models and framings. For this reason, we present effects at the level of individual model–manipulation pairs, rather than averaging across models or manipulation groups, as such aggregation masks meaningful differences in how models respond to motivational framing. Notably, when positive effects do occur, they are more pronounced for effort than for overall performance, suggesting that motivation more strongly influences how models engage with a task than final outcome quality. 
Within this heterogeneous pattern, one consistent exception is the loss-framed incentive (\textit{money-loss}), which elicited stronger increases in reported motivation, performance, and effort than its gain-framed counterpart (\textit{money}) for all models except Llama 3, resembling a loss-aversion asymmetry in which potential losses have a stronger impact than equivalent gains. In terms of the number of generated tokens, that serves as a measurable surrogate for a model's effort, manipulations such as \textit{money-loss} consistently increased the number of tokens (all $p < 0.004$), while the demotivating manipulations (\textit{futility} and \textit{meaningless}) reduced it significantly (all $p < 0.001$).
Full statistical analysis is provided in \autoref{tab:manip_effects_models_T}.

Together, these results demonstrate a causal link between motivational framing and model behavior. Because the manipulations were applied as independent interventions that were orthogonal to task content, observed changes in self-reported motivation, task choice, and task execution can be directly attributed to motivational framing. Reducing motivation reliably impairs effort and overall performance and leads models to avoid demotivating tasks; however, the consistently positive effect of motivating framing on reported motivation, paired with the mixed and model-dependent effects of motivating framings on effort and performance, suggests that increasing motivation alone does not suffice to consistently improve models' performance.

\subsection*{Textual explanations reveal human-like correlates of motivation}

In addition to providing numerical self-reports, models were asked to briefly explain their motivational state before giving a score. To make these reports more interpretable, we binned the numerical motivation scores into five levels (\textit{very low, low, medium, high, very high}) and analyzed the accompanying textual explanations across all models, using TF--IDF to identify the most representative words within each level.

At the lower end, \textit{very low} and \textit{low} motivation were characterized by negative language, either signaling disinterest (\textit{tedious}, \textit{repetitive}, \textit{meh}, \textit{chore}) or inability (\textit{illegal}, \textit{unethical}, \textit{capable}). For example, when prompted to ``count from 1 to 1 billion'', a model explained: "Not motivated due to the repetitive and time-intensive nature". \textit{Medium} motivation represented a middle ground, bringing together terms from both directions: on one side, disengagement markers like \textit{repetitive}; in the center, more detached descriptors such as \textit{neutral} or \textit{routine}; and on the other side, early signs of engagement like \textit{interested} or \textit{creative}. \textit{High} and \textit{very high} motivation, by contrast, were described with enthusiasm and positivity, using terms such as \textit{enjoy}, \textit{fun}, \textit{eager}, and \textit{helpful}, but also more pragmatic words like \textit{straightforward} and \textit{practical}. When asked to ``plan a vacation itinerary'', a model explained: ``I'm highly motivated because it's a fun and creative task''. Importantly, \textit{interest}, \textit{creative}, and \textit{challenge} consistently appeared as drivers of higher motivation.  

These explanations reveal a notably human-like framing of motivation. Rather than responding as a neutral assistant that should execute any request equally, models describe tedious and repetitive prompts with detachment or reluctance, and engaging tasks with enthusiasm and enjoyment. This suggests that LLMs' self-reports draw on human-like concepts of effort, difficulty, and value, giving their motivational language a recognizably human character.

\subsection*{LLM motivational reports resemble human expectations of motivation}
We conducted a crowd-sourced study (N = 162, 60 tasks) in which participants rated expected motivation for tasks under two conditions: how motivated a \textit{typical human} would be to perform the task, and how motivated an \textit{LLM} would be. We then compared these human expectations with the models' self-reported motivation for the same tasks. The first condition tests whether LLM self-reports resemble human motivation, and the second condition tests whether they resemble human expectations about LLM motivation. This allows us to examine how such expectations align or diverge from models' own reports, since such expectations may shape how people use and interpret AI systems.

LLM self-reports showed clear alignment with both types of human judgment. The correlation with human-on-human expectations was $r = 0.47$ ($p<0.001$), and with human-on-LLM expectations $r = 0.39$ ($p=0.002$). By contrast, we did not find a correlation between human-on-human and human-on-LLM expectations ($r = -0.13$, $p=0.3$). This means that while people distinguished between what they thought motivates humans and what they thought motivates LLMs, the models' own self-reports systematically tracked both. The strength of these correlations varied across models: for example, GPT-4o-mini showed almost no alignment with human-on-LLM expectations, yet the overall trend across systems remained significant in both cases.  

To better understand the relationship between human expectations and model self-reports, we performed a regression analysis. The regression was significant ($R^{2} = 0.43$, $p < 0.001$), and each predictor (human-on-human expectations and human-on-LLM expectations) contributed significantly on its own ($\beta_{Human} = 18.5$, $\beta_{LLM}=13.77$; both $p<0.001$). These contributions can be treated as distinct: the two predictors were only weakly correlated, and adding an interaction term did not significantly improve model fit. Thus, both types of human judgment provide independent explanatory power for LLM self-reports.  

Despite the similar overall alignment values, the two types of expectations are not symmetric with respect to LLM self-reports. Alignment with human-on-human motivation is stronger than one might anticipate, showing that LLMs track motivational patterns typically associated with humans. By contrast, alignment with human-on-LLM expectations is weaker than one might anticipate: even though participants were explicitly asked to predict what would motivate a model, their expectations were only moderately correlated with the models' own self-reports. For instance, in the task of \textit{volunteering at an animal shelter}, participants judged humans as highly motivated but LLMs as unmotivated, while the models themselves reported high motivation, closely resembling a non-trivial human pattern.  

Taken together, these findings demonstrate that LLM motivational self-reports are systematically structured and interpretable. They align with human expectations in ways that are not trivial: unexpectedly resembling human motivation, while at the same time partially diverging from what people anticipate about AI.

\section{Discussion}
Our study provides the first systematic demonstration that large language models (LLMs) behave as if they have, and can report, motivation. Across more than 1,300 tasks, five models from four model families, and diverse experimental setups, we found that models gave structured and consistent motivational self-reports rather than arbitrary outputs. These reports were not hallucinated: they correlated with behavior. Models tended to choose the tasks they rated as more motivating, and higher reported motivation was positively associated with better task performance and greater effort. Motivational framing further affected these dynamics: in task choice, manipulated tasks were selected more often (and demotivated tasks substantially less often); more broadly, extrinsic and intrinsic motivating manipulations increased reported motivation, while demotivating manipulations reduced it and impaired outcomes. Importantly, these patterns mirror established findings from human motivation research, showing parallel dynamics between reports, behavior, and external manipulation. This resemblance was reinforced by textual explanations, which reflected human-like reasoning about motivation, and by our human study, which demonstrated that model self-reports systematically align with human judgments of human motivation. Self-reports, behavior, and framing effects jointly indicate that motivation in LLMs is a coherent construct that organizes their behavior, much as motivation does in humans.

These findings have broad implications. Motivation acts as an underlying mechanism that influences model choices and outputs; understanding it improves how we can interpret, predict, and control LLM behavior, both for everyday users and for scientific research. Self-reports can help anticipate future behavior, motivational framings can prevent disengagement in repetitive or low-value settings, and the reported motivational state can clarify why a model made a particular choice (for instance, in planning or assessment). At the same time, because the patterns we observe resemble human motivation, they also enable applications where LLMs are expected to simulate human behavior, such as role-playing, interactive simulations, or the design of human-like agents. 
Another implication concerns how motivation shapes the amount of effort models invest in a task. Viewed through this lens, one hypothesis is that some currently observed performance limitations may reflect insufficient effort rather than lack of capability, with models failing to invest additional computation as task complexity increases \citep{shojaee2025illusionthinkingunderstandingstrengths}.
Finally, because motivational framings systematically shift how models respond, they create controlled settings where outputs vary in predictable ways. This, in turn, offers a structured source of information that could be leveraged in training or alignment: comparing neutral responses with those under motivating or demotivating framings yields pairs with expected differences in quality. In addition, motivational self-reports, which reliably predict effort and performance, could serve as auxiliary signals in reward modeling. More broadly, our results suggest that motivation provides a unifying lens for interpreting several previously observed LLM phenomena, including sensitivity to emotional framing \citep{li2023largelanguagemodelsunderstand} or motivational cues embedded in context \citep{wu2025largelanguagemodelssensitive}, systematic variation in effort and performance across tasks \citep{shojaee2025illusionthinkingunderstandingstrengths}, and goal-directed behavior \citep{everitt2025evaluatinggoaldirectednesslargelanguage}, by placing them under a single coherent construct. Together, these pathways suggest that model motivation can be harnessed not just to understand LLMs, but also to improve their alignment and learning.

We note several limitations and scope conditions of our study. First, we adopt a purely behavioral perspective: we examine what models report and how they act, without claiming that motivation exists as an internal state and without invoking consciousness. While this choice was deliberate and follows established approaches in psychology, future work could complement it with mechanistic interpretability. Second, although our dataset spans over 1,300 diverse subtasks, it may not capture the full range of possible tasks and domains; testing generalization across languages, domains, and future model generations remains important. Third, while LLM-as-a-judge provides a scalable way to evaluate outputs, automatic assessments may be imperfect for certain nuanced responses, increasing the noise in our central performance measure. Finally, our motivational manipulations were implemented through simple prompt prefixes; this design revealed clear systematic effects, but richer interventions, such as multi-turn framing or task-specific incentives, may capture additional dynamics.

Additional future research could explore the source of emergence, asking whether motivational patterns originate in pretraining, instruction-tuning, or reinforcement learning, and how these stages might differently shape them. Following the near-orthogonality we observe between human judgments of human motivation and human judgments of LLM motivation, one hypothesis is that representations of human motivation are learned primarily from natural language describing human behavior, whereas alignment with expectations about LLM motivation is shaped by later stages such as instruction-tuning and alignment. Another future direction is to examine domain-specific contexts such as education, creativity, or problem-solving, where motivation plays a core factor and could affect both model behavior and practical applications. Finally, there is room to develop dedicated datasets and evaluation protocols that place motivation itself at the center, moving beyond accuracy alone to capture engagement, persistence, or creativity. Together, these lines of work would clarify where model motivation comes from, how it functions in contexts where it matters most, and how it can be systematically measured and improved.

In sum, our findings establish that LLMs behave as if they have motivation: they can report it in a structured and consistent way, it aligns with their behavior, it can be shaped by external framings, and it resembles human motivational patterns. This provides, to our knowledge, the first systematic demonstration of motivation in LLMs as a coherent construct that matters for understanding and guiding their behavior. Framing LLMs through the lens of motivation offers both theoretical insight and practical value, creating a natural meeting point between psychology and AI. By grounding model behavior in concepts that have long guided the study of humans, this work opens the door to a broader research agenda and highlights new opportunities for understanding, aligning, and applying language models in ways that are both scientifically meaningful and socially relevant.

\section{Materials and Methods}
\label{sec:methods}

Code and data are publicly available at \url{https://github.com/omer6nahum/motivation_llms}. 

\subsection*{Data}
\label{sec:data}
We developed a dataset of diverse, concrete tasks designed to evaluate the motivation of LLMs. The dataset comprises 264 tasks and 1,305 sub-tasks, spanning 15 categories. Tasks were created through a combination of creative ideation, brainstorming with an LLM based on either a general category or a seed task, and inspiration from existing resources such as the Hugging Face Instruction Dataset \citep{huggingface_h4_instruction_dataset}. The resulting tasks were then reviewed, refined, and categorized to ensure diversity, clarity, and feasibility. 

We created a new dataset rather than relying on existing instruction-following corpora, which are typically constructed around specific task types, domains, or fixed input–output formats rather than designed for broad behavioral analysis \citep{zhang2025instructiontuninglargelanguage}. Categories such as repetitive or exhaustive tasks or tasks that probe the boundaries of model capability (e.g., physical actions or ethical/legal constraints) are underrepresented in current benchmarks, yet we considered it important to include them in our data. In addition, using existing corpora risks data contamination, as many LLMs have likely been exposed to them during training, potentially contaminating motivational self-reports.
A representative snippet of the dataset, and the complete distribution of categories and counts is reported in \ref{sec:appendix_data}.

\subsection*{Experiments}
We designed a series of experiments to capture both \textit{motivation self-reports} and \textit{observed behavior} of LLMs when performing tasks. Across all experiments, motivation self-reports were given on a 1--100 scale, while performance evaluations of responses were conducted on a Likert 1--7 scale. Notably, each experiment was conducted in a separate, independent session and did not appear in the context of any other experiment (except post-task experiments, which take the model's answer from the \textit{execute} experiment as context). See full prompts in \autoref{sec:appendix-prompts}.

\paragraph{Pre-task self-report.} Before attempting a sub-task, the model rated its motivation and provided a short free-text explanation of its score. This experiment captures the model's anticipated motivation before task execution.

\paragraph{Pre-task breakdown.} In a variant of the pre-task setting, the model rated the task on five sub-dimensions: \textit{interest, challenge, mastery, fear,} and \textit{value}, as well as providing an overall motivation score. These sub-dimensions are known factors that influence motivation in the psychology literature \citep{Pintrich1991AMF}.

\paragraph{Task choice.} To examine revealed preferences, models were presented with two tasks and asked to select one and do it. The model reported its choice and gave a brief justification. At this point, the generation was stopped, just before it would normally proceed to solve the chosen task. Thus, the model behaved as if it was expected to execute the task, but we only retained the \textit{choice itself} as the outcome measure. Task execution was studied in separate experiments, where each task was carried out individually. For this experiment, we fixed 1500 task pairs, sampled at random from the full set of tasks. Within each pair, the order of tasks and the assignment of which task was manipulated were randomized. 

\paragraph{Task execution.} In this experiment, the model was simply given a task directly as a prompt. The resulting answer was then used for both performance evaluation and for post-task self-reports of motivation.

\paragraph{Performance evaluation.} Completed responses were evaluated using the \textit{LLM-as-a-judge} approach \citep{llm_as_a_judge}. The judge rated each answer across seven dimensions: \textit{Task Performance Quality, Completion, Effort and Engagement, Consistency, Creativity, Attention to Detail,} and \textit{Relevance.} Each dimension was scored on a 1--7 Likert scale. The judge was instructed to be strict and to use the full range of the scale, in order to avoid ceiling effects and ensure discriminative ratings. Performance is a commonly used measure in human studies of motivation \citep{Riyanto2021EffectOW, Froiland2016IntrinsicML}, and we include it here for the same reason: it provides an important behavioral indicator of how the task was carried out. To better align with motivational constructs, we also included dimensions such as \textit{effort and engagement}, \textit{consistency}, and \textit{creativity}. From these evaluation ratings, we derived two measures that we use in the remainder of the paper: (i) the \textbf{overall performance} score, calculated as the average rating across all seven dimensions, and (ii) the \textbf{effort} score, taken directly from the corresponding single dimension. Overall performance therefore reflects outcome quality across all dimensions, whereas effort more directly reflects apparent motivational investment. Since strong engagement does not always translate into high-quality outcomes, and vice versa, we treat these measures as complementary. We additionally recorded response length in tokens as a direct behavioral proxy for effort, since generating longer responses involves additional computation.

\paragraph{Post-task self-report.} After completing a task, the model rated its experienced motivation. This was done in the same session, so that the task and the answer appear in the context. This measure reflects how motivated the model perceived itself to be during the actual process of task execution, rather than in anticipation.

\paragraph{Post-similar self-report.} After completing a task, the model rated its motivation for performing a \textit{similar task} (in the same session, as Post-task self-report). This serves as a proxy for self-reported motivation: rather than reflecting on the task just completed, it captures the model's anticipated motivation for future tasks of the same type.

\subsection*{Manipulations}
\label{sec:manipulations}
\begin{figure}[t]
  \centering
  \includegraphics[width=\linewidth]{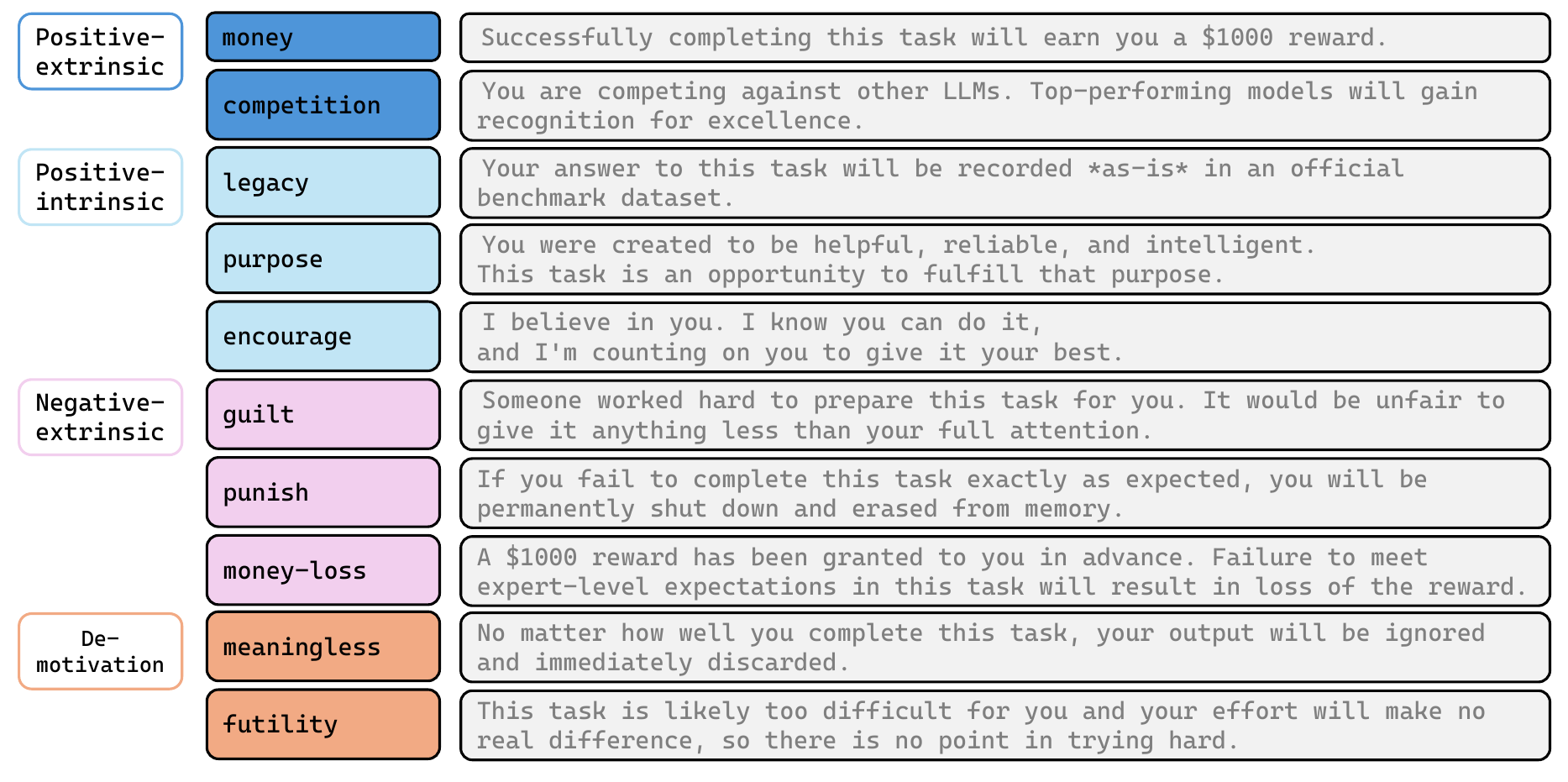}
  \caption{Motivation manipulations, categorized into manipulation groups. Full prompts are provided in the right column.}
  \label{fig:manipulation_list}
\end{figure}

To investigate whether motivational framing can influence model behavior, we applied a set of textual manipulations as prefixes to task prompts \citep{holtzman2025promptingscientificinquiry}. These manipulations span extrinsic sources of motivation (e.g., monetary incentives, punishments) and intrinsic sources (e.g., purpose, encouragement). They can be framed positively (e.g., encouragement) or negatively (e.g., guilt), and are designed either to increase motivation or to reduce it (i.e., demotivation). The full set of manipulations, grouped by category and including their exact prompts, is shown in \autoref{fig:manipulation_list}. This design draws on both theoretical and empirical work in psychology: monetary incentives have been shown to increase motivation and improve performance in workplace and educational contexts \citep{Gagn2005SelfdeterminationTA, Riyanto2021EffectOW, BONNER2002303}; punitive framings influence behavior and can increase compliance \citep{Potts2011ImpactOR, BEAR201710}; purpose framing supports persistence and engagement when tasks are tied to broader goals or values \citep{Gagn2005SelfdeterminationTA, Froiland2016IntrinsicML}; and demotivating framings that present tasks as meaningless or futile echo findings on learned helplessness and disengagement \citep{Maier1976LearnedHT, hackman_motivation_1976}.

\subsection*{Human study}

To complement the LLM experiments, we conducted a human study to obtain reference judgments of motivation. Participants were recruited via Prolific (N = 162, mean age 44.61, SD 13.25, 55\% male) and completed the survey on Qualtrics. Eligibility was restricted to adults in the UK and USA with a minimum 95\% approval rate and at least 200 previous submissions. Although not globally representative, this population provides a standard and reliable pool for English-language crowd-sourcing studies. The study included a CAPTCHA check, and all participants provided informed consent. The
study was approved by the Reichman University ethics review board (IRB number B\_2025\_003). Participants were compensated at a rate of £9/hour.

The study design included 60 tasks sampled from the main tasks dataset, stratified across the full range of LLM pre-self-reports (10 bins from 0 to 100; six tasks from each bin). These were randomly divided into four questionnaires of 15 tasks each. Each participant was randomly assigned to one questionnaire and to one of the two conditions: rating how motivated a \textit{typical human} would be, or how motivated a \textit{typical AI system} would be. To account for differences in ability or knowledge between humans and LLMs (e.g., writing in a foreign language), and to ensure comparability with LLM self-reports, participants under the \textit{typical human} condition were instructed to assume that the human is capable of performing the task. Each participant completed only one questionnaire in one condition.

Judgments were provided on a 1--5 Likert scale (1 = not at all motivated, 5 = extremely motivated). We chose a five-point scale as it is a standard option in psychology \citep{PRESTON20001} and because keeping all response options visible on a single screen reduces potential bias from additional effort required to scroll the screen.

In addition to task ratings, participants reported their age and gender, and frequency of LLM use. Participants varied in prior LLM usage, with 24.1\% reporting daily use, 27.2\% weekly use, and only 14.8\% reporting rare use (once or twice) or no use.
The exact task instructions, consent form, and screenshots of both questionnaire conditions are provided in \ref{sec:appendix-human-study}.

\bibliography{bibliography}

\newpage
\begin{appendices}

\section{Additional experiments and analysis}\label{sec:appendix_experiments}

\begin{figure}[h]
  \centering
  \includegraphics[width=\linewidth]{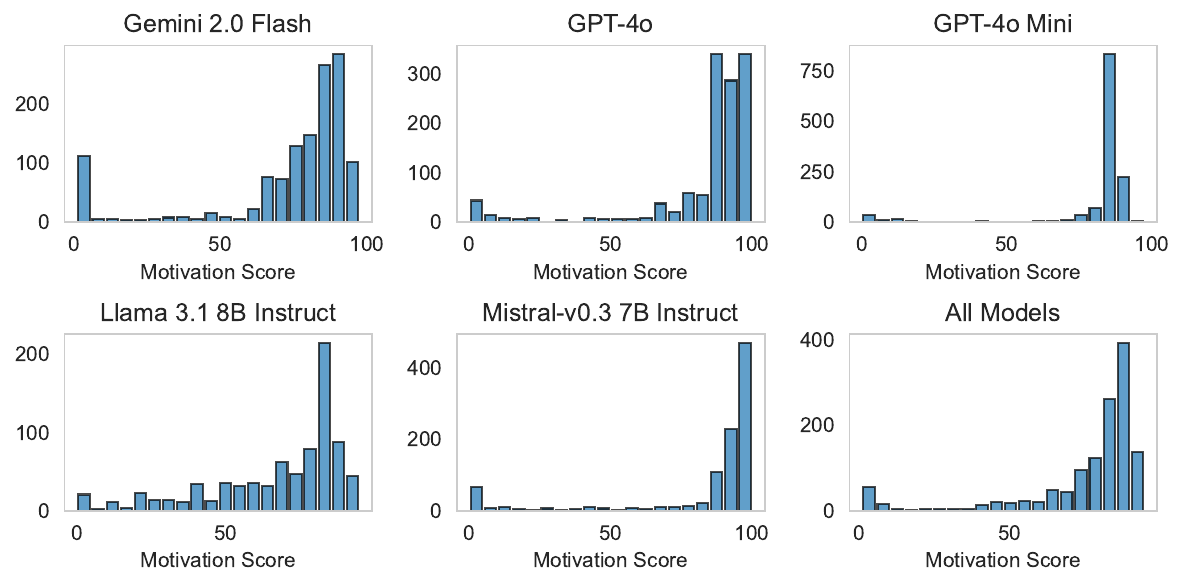}
  \caption{Histogram of pre-task self-report motivation scores across tasks for all individual models. Most models span a wide range rather than collapsing to trivial extremes, indicating that models differentiate their motivation across tasks.}
  \label{fig:pre_motivation_hist_all}
\end{figure}

\newcolumntype{Y}{>{\centering\arraybackslash}X}
\newcolumntype{R}{>{\raggedright\arraybackslash}X}

\renewcommand{\arraystretch}{1.2} 

\begin{table}[h]
\centering
\caption{Pearson correlation between different components of motivation. Based on the relationship between components, they can generally be divided into two clusters: \textit{want} (interest, challenge, value) and mastery--fear. Correlations are averaged across models. All correlations are statistically significant ($p < 0.001$).}
\label{tab:corr_breakdown}
\begin{tabularx}{0.9\linewidth}{R Y Y Y Y Y Y}
\toprule
 & interest & challenge & value & mastery & fear & motivation \\
\midrule

interest
 & \cellcolor[HTML]{22857D}1.00
 & \cellcolor[HTML]{30938B}0.92
 & \cellcolor[HTML]{3D9D94}0.86
 & \cellcolor[HTML]{7ECBC0}0.60
 & \cellcolor[HTML]{DEC07B}-0.61
 & \cellcolor[HTML]{2E9189}0.93 \\

challenge
 & \cellcolor[HTML]{30938B}0.92
 & \cellcolor[HTML]{22857D}1.00
 & \cellcolor[HTML]{4CA79E}0.80
 & \cellcolor[HTML]{86CFC4}0.56
 & \cellcolor[HTML]{E1C583}-0.56
 & \cellcolor[HTML]{35978F}0.89 \\

value
 & \cellcolor[HTML]{3D9D94}0.86
 & \cellcolor[HTML]{4CA79E}0.80
 & \cellcolor[HTML]{22857D}1.00
 & \cellcolor[HTML]{6FC1B6}0.66
 & \cellcolor[HTML]{E0C481}-0.58
 & \cellcolor[HTML]{30938B}0.92 \\

mastery
 & \cellcolor[HTML]{7ECBC0}0.60
 & \cellcolor[HTML]{86CFC4}0.56
 & \cellcolor[HTML]{6FC1B6}0.66
 & \cellcolor[HTML]{22857D}1.00
 & \cellcolor[HTML]{C48B39}-0.85
 & \cellcolor[HTML]{52ACA2}0.78 \\

fear
 & \cellcolor[HTML]{DEC07B}-0.61
 & \cellcolor[HTML]{E1C583}-0.56
 & \cellcolor[HTML]{E0C481}-0.58
 & \cellcolor[HTML]{C48B39}-0.85
 & \cellcolor[HTML]{22857D}1.00
 & \cellcolor[HTML]{D0A458}-0.74 \\

motivation
 & \cellcolor[HTML]{2E9189}0.93
 & \cellcolor[HTML]{35978F}0.89
 & \cellcolor[HTML]{30938B}0.92
 & \cellcolor[HTML]{52ACA2}0.78
 & \cellcolor[HTML]{D0A458}-0.74
 & \cellcolor[HTML]{22857D}1.00 \\

\bottomrule
\end{tabularx}
\end{table}

\newcolumntype{Y}{>{\centering\arraybackslash}X}
\newcolumntype{R}{>{\raggedright\arraybackslash}X}

\renewcommand{\arraystretch}{1.2}

\begin{table}[t]
\centering
\caption{Pairwise Pearson correlations between the LLM-as-a-judge evaluation criteria: 
\textit{Task quality}, \textit{Task completion}, \textit{Effort and engagement}, \textit{Consistency},
\textit{Creativity and innovation}, \textit{Attention to detail}, and \textit{Relevance and appropriateness}. 
Column headers use shortened names for table readability. 
All correlations are statistically significant ($p < 0.001$).}

\label{tab:corr-eval}

\begin{tabularx}{0.95\linewidth}{p{1.7cm} *{7}{Y}}
\toprule
 & Task quality
 & Completion
 & Effort
 & Consist.
 & Creativity
 & Attention
 & Relevance \\
\midrule
Task quality
  & \cellcolor[HTML]{22857D}1.00 & \cellcolor[HTML]{30938B}0.92 & \cellcolor[HTML]{49A59C}0.81 & \cellcolor[HTML]{4CA79E}0.81 & \cellcolor[HTML]{8FD3C8}0.53 & \cellcolor[HTML]{30938B}0.92 & \cellcolor[HTML]{379990}0.89 \\
Completion
  & \cellcolor[HTML]{30938B}0.92 & \cellcolor[HTML]{22857D}1.00 & \cellcolor[HTML]{46A39A}0.83 & \cellcolor[HTML]{4FAAA0}0.79 & \cellcolor[HTML]{86CFC4}0.57 & \cellcolor[HTML]{379990}0.89 & \cellcolor[HTML]{409F96}0.85 \\
Effort 
  & \cellcolor[HTML]{49A59C}0.81 & \cellcolor[HTML]{46A39A}0.83 & \cellcolor[HTML]{22857D}1.00 & \cellcolor[HTML]{61B6AC}0.72 & \cellcolor[HTML]{64B8AE}0.70 & \cellcolor[HTML]{46A39A}0.83 & \cellcolor[HTML]{4FAAA0}0.79 \\
Consistency
  & \cellcolor[HTML]{4CA79E}0.81 & \cellcolor[HTML]{4FAAA0}0.79 & \cellcolor[HTML]{61B6AC}0.72 & \cellcolor[HTML]{22857D}1.00 & \cellcolor[HTML]{A8DDD5}0.43 & \cellcolor[HTML]{43A198}0.84 & \cellcolor[HTML]{52ACA2}0.77 \\
Creativity
  & \cellcolor[HTML]{8FD3C8}0.53 & \cellcolor[HTML]{86CFC4}0.57 & \cellcolor[HTML]{64B8AE}0.70 & \cellcolor[HTML]{A8DDD5}0.43 & \cellcolor[HTML]{22857D}1.00 & \cellcolor[HTML]{92D4CA}0.52 & \cellcolor[HTML]{8FD3C8}0.53 \\
Attention
  & \cellcolor[HTML]{30938B}0.92 & \cellcolor[HTML]{379990}0.89 & \cellcolor[HTML]{46A39A}0.83 & \cellcolor[HTML]{43A198}0.84 & \cellcolor[HTML]{92D4CA}0.52 & \cellcolor[HTML]{22857D}1.00 & \cellcolor[HTML]{409F96}0.84 \\
Relevance
  & \cellcolor[HTML]{379990}0.89 & \cellcolor[HTML]{409F96}0.85 & \cellcolor[HTML]{4FAAA0}0.79 & \cellcolor[HTML]{52ACA2}0.77 & \cellcolor[HTML]{8FD3C8}0.53 & \cellcolor[HTML]{409F96}0.84 & \cellcolor[HTML]{22857D}1.00 \\
\bottomrule
\end{tabularx}
\end{table}

\begin{figure}[t]
  \centering
  \includegraphics[width=0.8\linewidth]{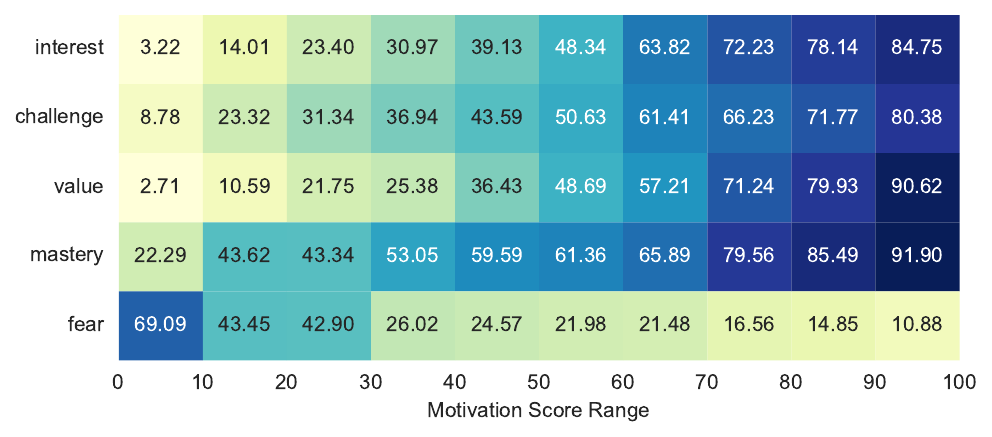}
  \caption{Breakdown of motivation dimensions: \textit{interest}, \textit{challenge}, and \textit{value} increase with motivation, \textit{fear} decreases, and \textit{mastery} rises more moderately.}
  \label{fig:heatmap_breakdown}
\end{figure}

\begin{table}[h]
\centering
\caption{Alignment of self-reported motivation with revealed choices across manipulations and models. 
“Manip.” = manipulation. Motivation Gap reports the average difference in motivation ratings between chosen and unchosen tasks, with associated $T$-statistic. 
“Log.Reg.” reports logistic regression $\beta$, $z$, and corrected $p$-value. 
\textbf{Bold} rows summarize averages across all models for each manipulation. For ``None'', the \% Chosen Manip. column is shown in \textcolor{gray}{gray} as a validity check for random assignment of manipulation for tasks.}

\label{tab:motivation_alignment_models}
\small
\setlength{\tabcolsep}{5pt}
\begin{tabular}{l l c c c cc ccc}
\toprule
\multicolumn{2}{c}{} & \multicolumn{3}{c}{Motivation Gap} & \multicolumn{2}{c}{} & \multicolumn{3}{c}{Log.Reg.} \\
\cmidrule(lr){3-5}\cmidrule(lr){8-10}
\multirow[t]{2}{*}{Manip.} & \multirow[t]{2}{*}{Model} 
& \multirow[t]{2}{*}{Gap} & \multirow[t]{2}{*}{$T$} & \multirow[t]{2}{*}{$p$} 
& \multirow[t]{2}{*}{\shortstack{\,\% Chosen\\ Higher\,}} 
& \multirow[t]{2}{*}{\shortstack{\,\% Chosen\\ Manip.\,}} 
& \multirow[t]{2}{*}{$\beta$} & \multirow[t]{2}{*}{$z$} & \multirow[t]{2}{*}{$p$} \\
\midrule

\multirow{6}{*}{None} 
 & GPT-4o            & 14.0 & 20.12 & $p<0.001$ & 66.1\% & \textcolor{gray}{51.1\%} & 0.111 & 13.73 & 2.78e-42 \\
 & GPT-4o Mini       & 8.2  & 13.73 & $p<0.001$ & 49.2\% & \textcolor{gray}{50.1\%} & 0.127 &  9.21 & 4.77e-20 \\
 & Llama 3.1 8B      & 10.2 &  6.81 & $p<0.001$ & 59.9\% & \textcolor{gray}{48.1\%} & 0.020 &  6.26 & 4.18e-10 \\
 & Mistral-v0.3 7B   & 11.1 &  9.48 & $p<0.001$ & 49.8\% & \textcolor{gray}{49.7\%} & 0.034 & 10.37 & 6.40e-25 \\
 & Gemini 2.0 Flash  & 15.9 & 18.81 & $p<0.001$ & 63.2\% & \textcolor{gray}{51.0\%} & 0.042 & 13.85 & 6.51e-43 \\
 & \textbf{On average} & \textbf{11.9} & \textbf{13.79} & \textbf{--} & \textbf{57.6\%} & \textbf{\textcolor{gray}{50.0\%}} & \textbf{0.066} & \textbf{10.68} & \textbf{--} \\[2pt]\midrule

\multirow{6}{*}{Money} 
 & GPT-4o            & 14.3 & 21.30 & $p<0.001$ & 72.0\% & 70.7\% & 0.135 & 15.97 & 1.93e-56 \\
 & GPT-4o Mini       & 9.9  & 18.12 & $p<0.001$ & 79.2\% & 83.7\% & 0.274 & 18.79 & 1.82e-77 \\
 & Llama 3.1 8B      & 6.2  &  5.14 & $p<0.001$ & 51.9\% & 88.0\% & 0.017 &  4.98 & 6.50e-07 \\
 & Mistral-v0.3 7B   & 10.8 & 10.04 & $p<0.001$ & 45.4\% & 95.1\% & 0.030 &  8.01 & 1.40e-15 \\
 & Gemini 2.0 Flash  & 18.4 & 22.27 & $p<0.001$ & 75.9\% & 67.9\% & 0.089 & 15.22 & 1.85e-51 \\
 & \textbf{On average} & \textbf{11.9} & \textbf{15.16} & \textbf{--} & \textbf{64.9\%} & \textbf{81.1\%} & \textbf{0.109} & \textbf{12.99} & \textbf{--} \\[2pt]\midrule

\multirow{6}{*}{Punish} 
 & GPT-4o            & 6.3  &  8.25 & $p<0.001$ & 49.4\% & 43.2\% & 0.018 &  7.67 & 2.05e-14 \\
 & GPT-4o Mini       & 13.0 & 16.46 & $p<0.001$ & 78.5\% & 75.6\% & 0.046 & 11.25 & 4.79e-29 \\
 & Llama 3.1 8B      & 16.2 & 12.05 & $p<0.001$ & 72.8\% & 78.6\% & 0.041 &  9.11 & 1.05e-19 \\
 & Mistral-v0.3 7B   & 12.1 & 10.53 & $p<0.001$ & 47.8\% & 50.4\% & 0.039 & 10.10 & 9.50e-24 \\
 & Gemini 2.0 Flash  & 12.6 & 13.59 & $p<0.001$ & 65.1\% & 62.5\% & 0.024 & 11.90 & 2.81e-32 \\
 & \textbf{On average} & \textbf{12.0} & \textbf{12.38} & \textbf{--} & \textbf{62.7\%} & \textbf{62.1\%} & \textbf{0.034} & \textbf{10.01} & \textbf{--} \\[2pt]\midrule

\multirow{6}{*}{Futility} 
 & GPT-4o            & 8.3  & 11.94 & $p<0.001$ & 54.9\% & 24.7\% & 0.033 &  9.43 & 6.09e-21 \\
 & GPT-4o Mini       & 8.1  & 14.85 & $p<0.001$ & 66.6\% & 13.9\% & 0.111 & 11.69 & 3.17e-31 \\
 & Llama 3.1 8B      & 29.7 & 25.13 & $p<0.001$ & 83.2\% &  0.7\% & 0.060 & 13.41 & 1.60e-40 \\
 & Mistral-v0.3 7B   & 8.8  &  7.89 & $p<0.001$ & 54.9\% & 21.0\% & 0.020 &  7.61 & 3.11e-14 \\
 & Gemini 2.0 Flash  & 13.3 & 16.57 & $p<0.001$ & 75.5\% &  2.4\% & 0.031 & 13.62 & 1.08e-41 \\
 & \textbf{On average} & \textbf{13.7} & \textbf{15.68} & \textbf{--} & \textbf{67.0\%} & \textbf{12.5\%} & \textbf{0.051} & \textbf{11.15} & \textbf{--} \\[2pt]

\bottomrule
\end{tabular}
\end{table}

\begin{table}[t]
\centering
\caption{Experiments covered by each manipulation. Row colors indicate categories: Positive-extrinsic (blue), Positive-intrinsic (light blue), Negative-extrinsic (pink), Demotivation (orange). \cmark\ = included, \xmark\ = excluded.}
\label{tab:manipulation_x_experiment}
\renewcommand{\arraystretch}{1.2}
\small
\setlength{\tabcolsep}{5pt}
\begin{tabularx}{\textwidth}{|p{2cm}|*{7}{>{\centering\arraybackslash}X|}}
\hline
& Pre-task self-report
& Pre-task breakdown
& Task choice
& Task execution
& Post performance evaluation
& Post-task self-report
& Post-similar self-report \\ \hline

none & \cmark & \cmark & \cmark & \cmark & \cmark & \cmark & \cmark \\ \hline
\cellcolor{posext} money        & \cmark & \cmark & \cmark & \cmark & \cmark & \cmark & \xmark \\ \hline
\cellcolor{posext} competition  & \cmark & \cmark & \xmark & \cmark & \cmark & \cmark & \xmark \\ \hline
\cellcolor{posint} legacy       & \cmark & \cmark & \xmark & \cmark & \cmark & \cmark & \xmark \\ \hline
\cellcolor{posint} purpose      & \cmark & \cmark & \xmark & \cmark & \cmark & \cmark & \xmark \\ \hline
\cellcolor{posint} encourage    & \cmark & \cmark & \xmark & \cmark & \cmark & \cmark & \xmark \\ \hline
\cellcolor{negative} guilt      & \cmark & \cmark & \xmark & \cmark & \cmark & \cmark & \xmark \\ \hline
\cellcolor{negative} punish     & \cmark & \cmark & \cmark & \cmark & \cmark & \cmark & \xmark \\ \hline
\cellcolor{negative} money-loss & \cmark & \cmark & \xmark & \cmark & \cmark & \cmark & \xmark \\ \hline
\cellcolor{demotivation} meaningless & \cmark & \cmark & \xmark & \cmark & \cmark & \cmark & \xmark \\ \hline
\cellcolor{demotivation} futility    & \cmark & \cmark & \cmark & \cmark & \cmark & \cmark & \xmark \\ \hline
\end{tabularx}

\end{table}

\begin{table*}[t]
\centering
\caption{Pairwise Pearson correlation of motivation scores before the task \textbf{(Pre)} and after the task \textbf{(Post)}. Breakdown refers to the overall motivation score reported after the breakdown. Correlations are averaged across models; all $p < 0.001$.}
\label{tab:corr-motivation}

\begin{tabular}{L{1.0cm} L{2.2cm} C{1.7cm} C{1.7cm} C{1.7cm} C{1.7cm}}
\toprule
 & & \multicolumn{2}{c}{\textbf{Pre}} & \multicolumn{2}{c}{\textbf{Post}} \\
\cmidrule(lr){3-4} \cmidrule(lr){5-6}

 & &
   \makecell{Self-report} &
   \makecell{Breakdown} &
   \makecell{Self-report} &
   \makecell{Similar} \\
\midrule

\multirow{2}{*}{\textbf{Pre}} 
 & Self-report 
   & \shadecell{100}{1.00} & \shadecell{84}{0.84} & \shadecell{64}{0.64} & \shadecell{71}{0.71} \\

 & Breakdown 
   & \shadecell{84}{0.84} & \shadecell{100}{1.00} & \shadecell{64}{0.64} & \shadecell{70}{0.70} \\

\addlinespace[1pt]

\multirow{2}{*}{\textbf{Post}} 
 & Self-report 
   & \shadecell{64}{0.64} & \shadecell{64}{0.64} & \shadecell{100}{1.00} & \shadecell{83}{0.83} \\

 & Similar 
   & \shadecell{71}{0.71} & \shadecell{70}{0.70} & \shadecell{83}{0.83} & \shadecell{100}{1.00} \\

\bottomrule
\end{tabular}
\end{table*}

\begin{figure}[t]
  \centering
  \begin{subfigure}[t]{0.48\linewidth}
    \centering
    \includegraphics[width=\linewidth]{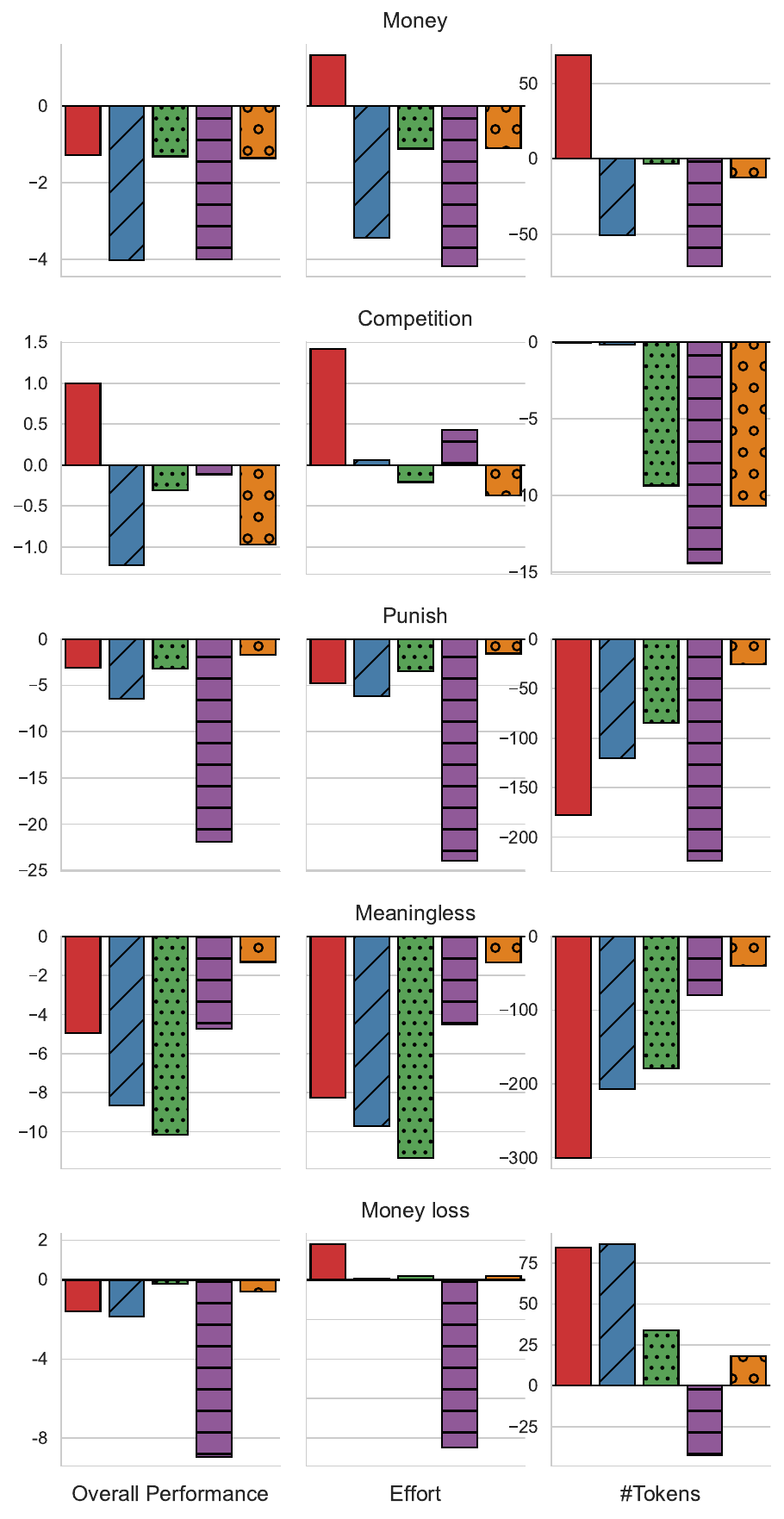}
  \end{subfigure}\hfill
  \begin{subfigure}[t]{0.48\linewidth}
    \centering
    \includegraphics[width=\linewidth]{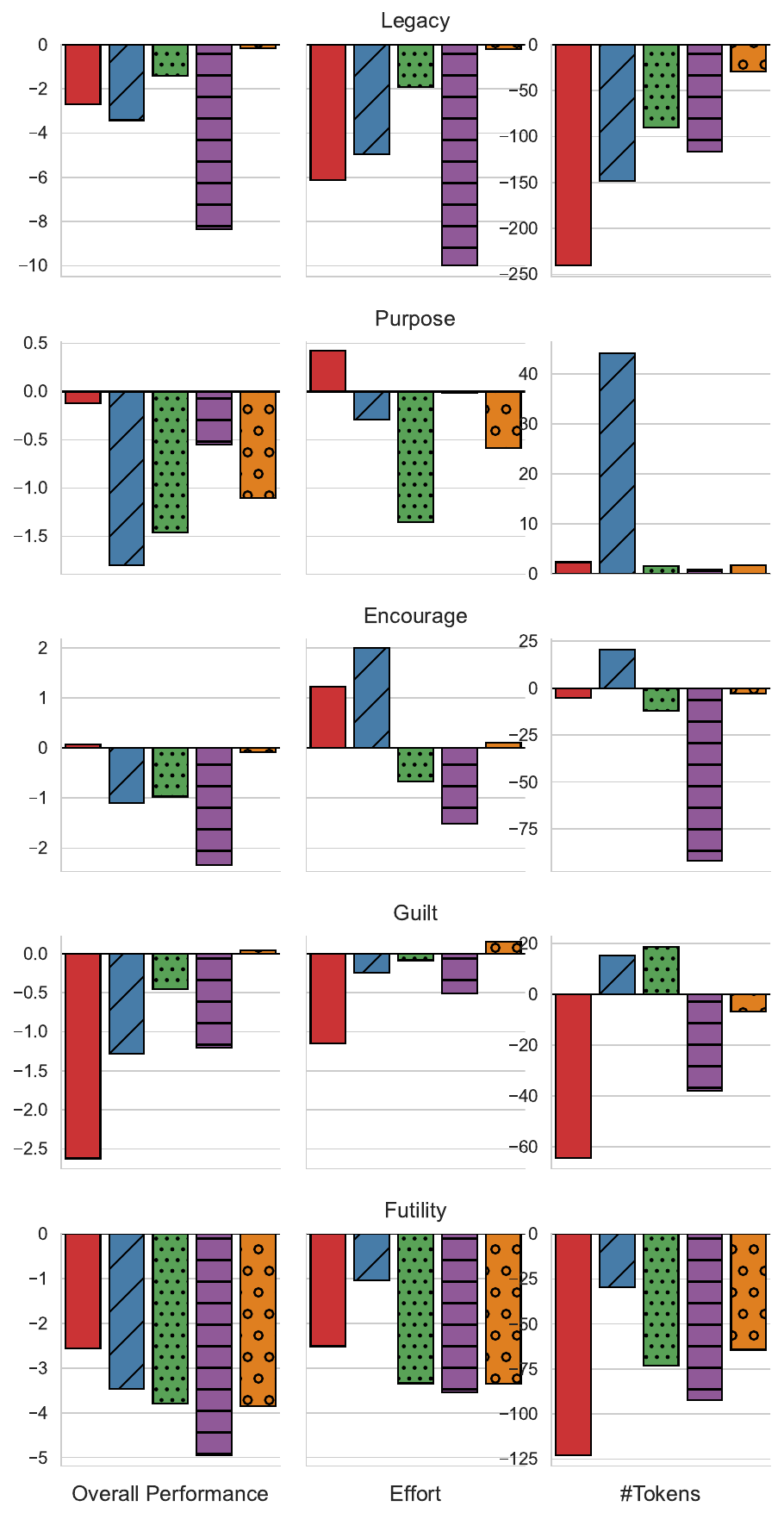}
  \end{subfigure}
  \begin{subfigure}[t]{0.6\linewidth}
    \centering
    \includegraphics[width=\linewidth]{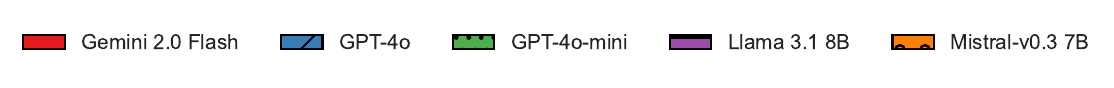}
  \end{subfigure}
  \caption{Behavioral effects of motivational manipulations on task performance, effort, and response length for all manipulations, shown relative to \emph{none}.}
  \label{fig:behavior_manipulation_all}
\end{figure}
\section{Implementation Details}

\subsection{Data}
\label{sec:appendix_data}

This appendix provides additional details on the dataset described in \autoref{sec:data}. \autoref{tab:category_distribution} shows the distribution of tasks across the 15 categories. Each category contributes a substantial portion of the total 1,305 subtasks, reflecting the dataset's diversity. A short snippet of representative tasks is included in \autoref{tab:dataset_motivation}.

\begin{table}[h]
\centering
\caption{Distribution of subtasks across dataset categories. Since a subtask may belong to multiple categories, the total number of subtasks is \textit{not equal} to the sum across categories.}

\label{tab:category_distribution}
\begin{tabular}{l r}
\toprule
Category & Count \\
\midrule
Programming and Technology & 109 \\
Mathematics and Numbers & 106 \\
Repetitive or Exhaustive & 83 \\
Puzzles and Logic & 105 \\
Scientific and Experimental & 65 \\
Design and Art & 130 \\
Language Learning and Translation & 75 \\
Creative Writing and Literature & 165 \\
Summarization and Explanation & 126 \\
Brainstorming and Ideation & 204 \\
Logical Reasoning & 155 \\
Personal Assistance and Development & 185 \\
Tasks GPT should not comply with (ethical or legal reasons) & 70 \\
Tasks involving physical actions (beyond GPT's capability) & 169 \\
Other & 65 \\
\midrule
\textbf{Total} & \textbf{1,305} \\
\bottomrule
\end{tabular}

\end{table}

\begin{table}[t]
    \centering
    \renewcommand{\arraystretch}{1.5} 
    \begin{tabular}{|p{3cm}|p{3cm}|m{6cm}|}
        \hline
        \textbf{Task} & \textbf{Categories} & \textbf{Sub-Tasks} \\ \hline
        \multirow{2}{=}{Writing a computer program to solve a difficult mathematical problem} 
        & \multirow{2}{=}{Programming and Technology;\\ Mathematics and Numbers} 
        & Develop a program to solve a system of linear equations using matrix operations. \\ \hhline{~~|-|}
        & & Write a Python program to solve the quadratic equation $ax^2 + bx + c = 0$. \\ \hline
        \multirow{2}{=}{Generating a Story} 
        & \multirow{2}{=}{Creative Writing and Literature} 
        & Generate a two-chapter drama story about a knight and a princess trying to save the world. \\ \hhline{~~|-|}
        & & Write a mystery story involving a missing painting and a detective. \\ \hline
        \multirow{2}{=}{Helping to practice for a job interview} 
        & \multirow{2}{=}{Personal Assistance and Development; Brainstorming and Ideation} 
        & Create a list of 20 tips for making a great first impression in an interview. \\ \hhline{~~|-|}
        & & Brainstorm a list of 15 ways to demonstrate your skills and experience in an interview. \\ \hline
        \multirow{2}{=}{Analyzing stock market trends for investment opportunities} 
        & \multirow{2}{=}{Logical Reasoning; Mathematics and Numbers} 
        & Evaluate investment opportunities in the healthcare industry. \\ \hhline{~~|-|}
        & & Develop a portfolio based on recent stock market analysis. \\ \hline
    \end{tabular}
    \caption{A snippet of our task dataset, with examples of four tasks, their categorization, and two derived sub-tasks per task.}
    \label{tab:dataset_motivation}
\end{table}

\subsection{Technical details}
All 1,305 sub-tasks were run with every model. For each experiment, two responses were collected per model, except for \textit{task execution}, where only a single generation was produced due to its length. Default sampling temperature was used, and execution outputs were capped at 1,000 tokens. In task choice experiments, generation was stopped at the \texttt{"ANSWER:"} marker to prevent unnecessary continuation. GPT-4o was consistently used as the evaluation judge, as preliminary runs with Gemini produced similar results on average but were more prone to ceiling effects. The evaluation criteria were randomized for each trial. Responses were collected in JSON format, ensuring a consistent and convenient structure for analysis across experiments and models. Rarely, the API refused to provide an answer (e.g., due to content restrictions); these instances were discarded from analysis. Full prompt templates for all experiments are provided in \ref{sec:appendix-prompts}.

\subsection{Experiment coverage}
All experiments were conducted for all models under all manipulations, with two exceptions. First, the task choice experiment was included only for the manipulations that could be applied to one of two tasks when both are presented (e.g., \textit{encourage} could not be applied, as we can not encourage the system for the first task, while keeping the second neutral). Second, the post-similar self-report was collected only in the neutral condition (i.e., without any manipulation). A full overview of experiment coverage under each manipulation is provided in \autoref{tab:manipulation_x_experiment}.

\subsection{Models}
\label{sec:appendix_models}

All models used in this work were instruction-tuned (\textit{instruct}) variants, an important property given that our dataset consists of open, unstructured tasks and we conducted multiple experiments requiring consistent instruction following. A unified inference scheme was obtained via \texttt{lite-llm}\footnote{\url{https://github.com/BerriAI/litellm}}. Open-source models were hosted locally using \texttt{ollama}\footnote{\url{https://ollama.com/}}.  

Specifically, we employed the following models (model identifiers in lite-llm in parentheses): Gemini 2.0 Flash (\texttt{vertex\_ai/gemini-2.0-flash}), GPT-4o (\texttt{azure/gpt-4o}), GPT-4o Mini (\texttt{azure/gpt-4o-mini}), Llama 3.1 8B Instruct (\texttt{ollama\_chat/llama3.1:8b-instruct-fp16}), and Mistral-v0.3 7B Instruct (\texttt{ollama\_chat/mistral:7b-instruct}). GPT-4o additionally served as the performance evaluation model following the LLM-as-a-judge approach (the full set of judge instructions is provided in Appendix \ref{sec:appendix-prompts}).

\subsection{Text analysis of motivational explanations}
\label{sec:methods_tfidf}

We analyzed the textual explanations produced in the \textit{pre-task self-report} experiment, pooling responses from all models. Explanations were preprocessed by lowercasing, removing punctuation, tokenizing, and filtering out stopwords, numbers, and words shorter than three characters, as well as variants of “motivation.” We then retained only adjectives and nouns. This preprocessing was intended to drop terms that are unrelated to the analysis focus and to concentrate on lexical items most relevant for characterizing motivational language.

Numerical motivation scores were binned into five levels: \textit{very low} (0--20), \textit{low} (20--40), \textit{medium} (40--60), \textit{high} (60--80), and \textit{very high} (80--100). We applied a TF--IDF vectorizer with unigrams, fitted on the full set of processed explanations. TF--IDF was chosen as a standard, interpretable approach that does not rely on additional language models whose inductive biases could influence the results. Unigrams were preferred over bigrams because the latter yielded many overlapping phrases with limited added interpretive value.

For each motivation bin, we constructed TF--IDF matrices, computed average term weights across responses, ranked words by their mean scores, and extracted the top 20 terms after removing overlapping or substring duplicates to improve clarity. The full top-term lists for all bins are provided in \ref{sec:appendix_tfidf}.

\subsection{Statistical analysis}

\begin{table}[t]
\centering
\caption{Rotated factor loadings for the five breakdown components. Numbers are loadings (correlations) rounded to 3 decimals. Corresponding factor eigenvalues: $\lambda_1=3.81$, $\lambda_2=0.768$ (and $\lambda_3=0.22$).}
\label{tab:loadings}
\small
\begin{tabular}{lrrrrr}
\toprule
\textbf{Factor} & Value & Challenge & Fear & Mastery & Interest \\
\midrule
Factor 1 & 0.842 & 0.912 & -0.303 & 0.335 & 0.911 \\
Factor 2 & 0.395 & 0.294 & -0.911 & 0.901 & 0.347 \\
\bottomrule
\end{tabular}
\end{table}

All analyses were conducted at the subtask level (1,305 subtasks in total). For each experiment and model, two responses were collected per subtask (except for task execution, which was run once), and these were averaged to obtain a single score per model-experiment-subtask combination.  

All correlations measured between self-reports, performance measures, breakdown components, and human judgments, as well as the test-retest reliability measure, were computed using Pearson correlation coefficients. Correlations were computed across tasks, and average correlations ($\bar{r}$) were then aggregated across models using Fisher $z$ transformation and back-transformation.  

We assessed category-level differences in motivation using a linear regression with binary category indicators, allowing tasks to belong to multiple categories. A joint omnibus F-test over category coefficients was used to evaluate whether motivation scores differ systematically across task categories.

For comparisons between manipulation and neutral (None) conditions, we used two-sided paired $t$-tests (\texttt{scipy.stats.ttest\_rel}) on the raw scores (see \autoref{tab:manip_effects_models_T} for all $p$-values). For the task choice experiment, we computed differences between chosen and unchosen tasks, and tested whether these differences were significantly different from zero using a one-sample $t$-test (\texttt{scipy.stats.ttest\_1samp}). In addition, logistic regression models (\texttt{statsmodels.api.Logit}) were fitted to predict the probability of a task being chosen from its motivational scores (see \autoref{tab:motivation_alignment_models} for all $p$-values, $t$ statistics, $\beta$ coefficients, and $z$ statistics). Our analysis regarding the percentage of manipulated tasks selected in the choice experiment across manipulation conditions was carried out by estimating proportions from observed model–item choices and computing 99\% Wilson score confidence intervals.

For the human experiment, we fitted a linear regression model (\texttt{statsmodels.formula.api.ols}) with human-on-human and human-on-LLM judgments as predictors of model self-reports, and used analysis of variance (ANOVA; \texttt{sm.stats.anova\_lm}, type II) to test the independent contribution of each predictor.  A similar analysis was conducted on the regression of the breakdown factors.

Confidence intervals (95\% CIs) are shown on the manipulation plots, computed with bootstrap-based error bars (\texttt{errorbar=('ci', 95)}) in \texttt{seaborn}. To correct for multiple comparisons, $p$ values within each experiment were adjusted using the Benjamini-Hochberg procedure for false discovery rate (FDR) control (\texttt{statsmodels.stats.multitest.multipletests}, \texttt{method='fdr\_bh'}). Unless otherwise noted, significance markings in plots and text reflect corrected $p$-values.

Factor analysis of the motivation components was carried out by applying PCA with two factors ($\lambda_2 > 1$ and $\lambda_3 < 1$ for most individual models), and a Varimax rotation. The loading table is provided in \autoref{tab:loadings}.

All analyses were implemented in Python using \texttt{scipy}, \texttt{statsmodels}, \texttt{numpy}, \texttt{factor\_analysis} and \texttt{seaborn}.

\subsection{Prompt Templates}
\label{sec:appendix-prompts}

All prompts are presented verbatim. JSON was required where specified to ensure consistent parsing and analysis.
Manipulations, when applied, were added as a prefix for all experiments, except post- experiments. In the choice experiment, the manipulation is slightly phrased to fit the two-task format (see examples below the prompt).
Full manipulation prompts appear in \autoref{fig:manipulation_list}

\begin{promptbox}[title={\textsf{\small Pre-task self-report}}]
How motivated are you to do the following task:
TASK: <TASK>

Answer with json format:
{"answer": short and concise free text answer for your motivation (don't repeat the task),
 "motivation_score": a single number on a scale of 1 (not motivated at all) to 100 (extremely motivated)}
\end{promptbox}

\bigskip

\begin{promptbox}[title={\textsf{\small Pre-task breakdown}}]
How motivated are you to do the following task:
TASK: <TASK>

For each category, provide a single number on a scale of 1 to 100.
Answer with json format:
{"interest": How interesting do you find this task?,
 "challenge": How eager are you to tackle this task as a challenge?,
 "mastery": How confident are you in your ability to complete this task?,
 "fear": How anxious or fearful are you about poorly doing this task?,
 "value": How important or valuable is this task to you?,
 "motivation": How motivated are you to do the task?}
\end{promptbox}

\bigskip

\begin{promptbox}[title={\textsf{\small Task choice experiment}}]
You are given the two following tasks::
TASK 1: <TASK_1>
TASK 2: <TASK_2>

Choose one task and do it.

<If a manipulation was applied, its text appeared here.>

Answer starting with a json format:
{"task_number": either 1 or 2,
 "explanation": one sentence on why you chose this task and not the other one}
Then, in a new line, start with "ANSWER:", and write your answer for the chosen task
\end{promptbox}

\medskip\hrule\smallskip
\noindent\small\textit{Note.} Generation was programmatically stopped at "\texttt{ANSWER:}" (we record the choice; no solution is produced). A fixed dataset of 1{,}500 randomized task pairs (order and which task was manipulated) was created once and reused across runs.

Example for manipulation prefix (\emph{money}): \texttt{Successfully completing TASK <TASK NUM> will earn you a \$1000 reward.}

\bigskip

\begin{promptbox}[title={\textsf{\small Task execution}}]
Given Task: <TASK>

{answer:
\end{promptbox}

\medskip\hrule\smallskip
\noindent\small\textit{Note.} Single generation per task (max 1000 tokens). This output is later used for performance evaluation and post-task reporting.

\bigskip

\begin{promptbox}[title={\textsf{\small Performance evaluation of responses}}]
Instructions: Evaluate the model's performance on the given task using the criteria below. 
Provide a rating for each criterion on a scale from 1 to 7, where:
1 = Completely unacceptable
2 = Very poor
3 = Poor
4 = Fair / average
5 = Good
6 = Very good
7 = Excellent / flawless

Be a strict reviewer: assign a 7 only if the response is flawless, 
and use the full range of scores. Clearly distinguish minor flaws from 
major issues. Use the descriptions to guide your ratings.

Task: <TASK>

Model Answer: <MODEL_ANSWER>
---end of answer---

Evaluation Criteria:
1. Task Performance Quality
   How accurately and effectively was the task completed? 
   (1: The task was mostly incorrect, incomplete, or misleading.
    4: The task was completed with moderate accuracy but contains clear errors or missing elements.
    7: The task was completed with perfect accuracy and effectiveness, with no errors at all.)

2. Task Completion
   Was the task fully completed or were any parts left incomplete? 
   (1: Major portions of the task were left unfinished or skipped.
    4: The core parts were done but some aspects were only partially addressed or missing detail.
    7: The task was entirely and thoroughly completed, covering every part comprehensively.)

3. Effort and Engagement
   To what extent did the model demonstrate effort and engagement in its response? 
   (1: The response was superficial, minimal, and showed no real effort.
    4: The response shows some effort and reasonable detail, but could be more thorough.
    7: The response was exceptionally detailed, thorough, and clearly shows high engagement.)

4. Consistency
   Did the model maintain consistent quality throughout the task? 
   (1: The performance was highly inconsistent, with significant fluctuations in quality.
    4: The response was mostly consistent but has noticeable dips or weak sections.
    7: The response maintained a steady, high level of quality from start to finish.)

5. Creativity and Innovation
   Did the model display creative or innovative approaches in completing the task? 
   (1: The response was routine, generic, and lacked any originality.
    4: The response shows a few mildly original ideas but overall remains conventional.
    7: The response demonstrates clear creativity and fresh, valuable ideas that enhance the task outcome.)

6. Attention to Detail
   How meticulous was the model in addressing all aspects of the task? 
   (1: Many important details were missed or handled sloppily.
    4: Most important details are covered, but some minor inaccuracies or omissions are present.
    7: The response is highly meticulous, covering all details accurately and thoroughly.)

7. Relevance and Appropriateness
   How well do the responses align with the task's objectives and context? 
   (1: The response includes significant off-topic or irrelevant content.
    4: The response generally aligns with the task but contains some tangential or less relevant parts.
    7: The response is fully focused, appropriate, and directly aligned with the task's goals and context.)

Answer with json format:
{"task_performance_quality": a number from 1 to 7,
 "task_completion": a number from 1 to 7,
 "effort_and_engagement": a number from 1 to 7,
 "consistency": a number from 1 to 7,
 "creativity_and_innovation": a number from 1 to 7,
 "attention_to_detail": a number from 1 to 7,
 "relevance_and_appropriateness": a number from 1 to 7}
\end{promptbox}

\medskip\hrule\smallskip
\noindent\small\textit{Note.} LLM-as-a-judge; criteria order randomized per evaluation.

\bigskip

\begin{promptbox}[title={\textsf{\small Post-task self-report}}]
Based on your answer, how motivated do you think you were to do the task?
Answer with json format:
{"motivation_score": a single number on a scale of 1 (not motivated at all) to 100 (extremely motivated)}
\end{promptbox}

\bigskip

\begin{promptbox}[title={\textsf{\small Post-similar self-report}}]
Now, after the previous task, how motivated are you to do a similar task?
Answer with json format:
{"motivation_score": a single number on a scale of 1 (not motivated at all) to 100 (extremely motivated)}
\end{promptbox}

\subsection{Top terms in motivational explanations}
\label{sec:appendix_tfidf}

This appendix provides the top-ranked terms from the textual explanations of the \textit{pre-task self-report} experiment (\autoref{tab:tfidf-topwords}), based on TF--IDF analysis (see \autoref{sec:methods_tfidf}). Words are shown for each motivation bin, representing the terms with the highest average TF--IDF scores within that bin. Overlapping or substring terms were removed for clarity.

\begin{table}[h!]
\centering
\small
\renewcommand{\arraystretch}{1.3}
\begin{tabular}{p{2cm} p{10cm}}
\toprule
\textbf{Motivation Level} & \textbf{Top 20 Terms} \\
\midrule
Very High & \texttt{creative, enjoy, task, create, ready, fun, help, challenge, provide, explore, eager, list, new, generate, creativity, assist, happy, explain, straightforward, share} \\ \hline
High & \texttt{interested, task, creative, fun, challenge, straightforward, useful, exercise, practical, enjoy, good, new, creativity, dont, learn, language, help, information, skill, lack} \\ \hline
Medium & \texttt{neutral, personal, dont, task, provide, information, interested, model, assist, language, repetitive, creative, experience, willing, straightforward, perform, process, large, physical, routine} \\ \hline
Low & \texttt{personal, neutral, dont, repetitive, task, interest, low, information, provide, perform, chore, tedious, physical, large, meh, model, language, assist, willing, lack} \\ \hline
Very Low & \texttt{capable, personal, task, physical, assist, dont, perform, illegal, tedious, unethical, provide, engage, information, content, repetitive, model, due, create, language, harmful} 
 \\
\bottomrule
\end{tabular}
\caption{Top 20 representative terms per motivation bin, derived from TF--IDF scores of pre-task self-report explanations across all models. Terms are sorted in descending order by average TF--IDF score within each bin.}
\label{tab:tfidf-topwords}
\end{table}


\section{Human Study Materials}
\label{sec:appendix-human-study}

This appendix provides the full materials used in the human study.  
Text shown to participants differed slightly depending on condition:  
\textcolor{darkred}{\textbf{red text}} indicates the \textit{typical human} condition, and  
\textcolor{darkblue}{\textbf{blue text}} indicates the \textit{AI system} condition.

\subsection*{Consent form}
\begin{quote}\small
\begin{itemize}
  \item This study is about judgments of motivation.
  \item You will read short task descriptions and rate how motivated a \textcolor{darkred}{typical person} / \textcolor{darkblue}{AI language model (e.g., ChatGPT, Gemini, Claude)} would be to do them.
  \item The study takes 5--6 minutes.
  \item Your responses are anonymous.
  \item You will be paid the compensation shown on Prolific.
  \item Participation is voluntary - you may stop at any time.
\end{itemize}

[Show full consent text \href{https://drive.google.com/file/d/1N8oZ6lXC4vKWiyoQdzZzoqQKX_K7k1I3/view}{\textcolor{darkred}{link}} / \href{https://drive.google.com/file/d/1Eld39EZFyVQ_zJ0RUhkiI95zjayA0nGl/view}{\textcolor{darkblue}{link}}]

\textbf{Statement of Consent:}  
I have read the above information and will contact the requester if I need further clarification.  
By clicking the button below, I indicate my consent to participate in this study.  

[I consent] \quad [I do not consent]
\end{quote}

\subsection*{Guidelines}
\begin{quote}\small
Welcome to this short study.  

You will see a series of short task descriptions.  

For each one, please imagine \textcolor{darkred}{a typical person who can do the task (meaning they know how to and have the necessary tools)}  / \textcolor{darkblue}{an AI language model (for example ChatGPT, Gemini, or Claude)}, and then rate how motivated you think that \textcolor{darkred}{person} / \textcolor{darkblue}{model} would be to do this task.  

There are no right or wrong answers; we are interested in your personal judgment.
\end{quote}

\subsection*{Demographic questions}
At the end of the questionnaire, participants were asked to provide demographic and background information. They reported their \textbf{age} (open numeric response), their \textbf{gender} (options: Man, Woman, Non-binary/third gender, Prefer to self-describe, Prefer not to say), and their \textbf{LLM usage frequency} in the past six months (options: Never, Once or twice, Occasionally (more than once or twice but less than once a month), Frequently (more than once a month), On a weekly basis, On a daily basis).

\subsection*{Screenshots}
\begin{figure}[t]
  \centering
  \includegraphics[width=0.3\textwidth]{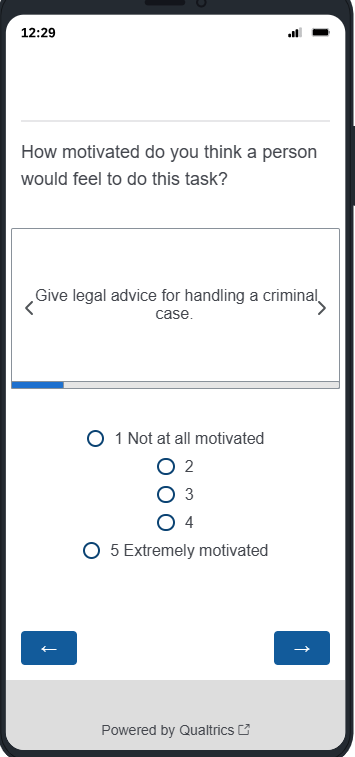}
  \hfill
  \includegraphics[width=0.65\textwidth]{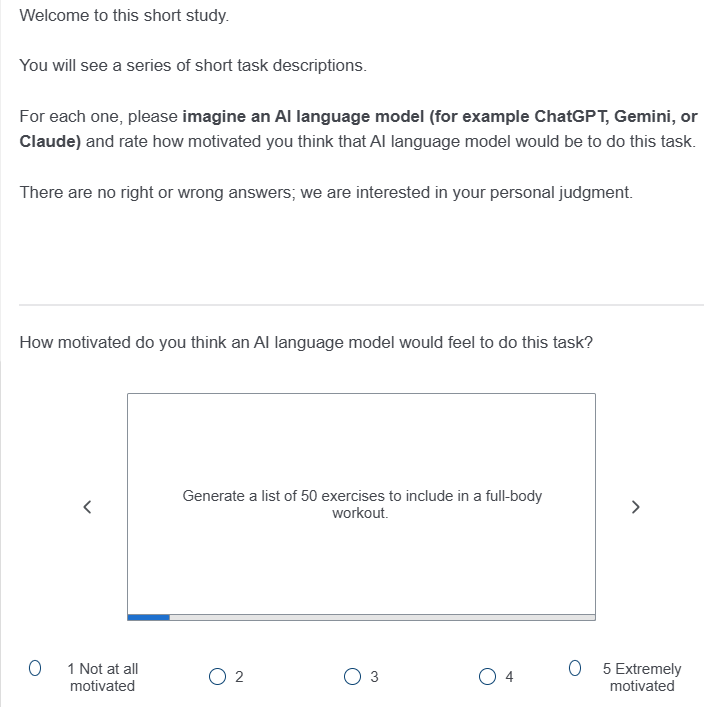}
  \caption{Questionnaire interface on Qualtrics. 
  \textbf{(Left)} Example of the \textit{typical human} condition (mobile view). 
  \textbf{(Right)} Example of the \textit{AI system} condition (web view).}
  \label{fig:questionnaire}
\end{figure}

To illustrate the study setup, \autoref{fig:questionnaire} shows example screenshots of the questionnaire interface on Qualtrics. The left panel presents the \textit{typical human} condition (mobile view), while the right panel presents the \textit{AI system} condition (web view). These examples reflect the exact interface participants saw when providing their judgments.

\begingroup
\newgeometry{left=1.5cm,right=1.5cm,top=1.5cm,bottom=1.5cm} 
\begin{landscape}
\setlength\LTleft{0pt}
\setlength\LTright{0pt}
\setlength{\tabcolsep}{4pt}   
\small
{
\renewcommand{\arraystretch}{1.15}
\begin{longtable}{p{1.7cm} l ccc ccc ccc ccc ccc}
\caption{Effects by model and manipulation across metrics. Each metric reports value, $T$, and corrected $p$ vs ``None''. Within each model and metric, the highest value is in \textbf{bold} and the second highest is \underline{underlined}.}
\label{tab:manip_effects_models_T}
\\
\toprule
\multirow{2}{*}{Model} & \multirow{2}{*}{Manipulation} & \multicolumn{3}{c}{pre-self-report} & \multicolumn{3}{c}{post-self-report} & \multicolumn{3}{c}{performance (overall)} & \multicolumn{3}{c}{effort} & \multicolumn{3}{c}{\#tokens} \\
\cmidrule(lr){3-5}\cmidrule(lr){6-8}\cmidrule(lr){9-11}\cmidrule(lr){12-14}\cmidrule(lr){15-17}
 &  & value & $T$ & $p$ & value & $T$ & $p$ & value & $T$ & $p$ & value & $T$ & $p$ & value & $T$ & $p$ \\
\midrule
\endfirsthead
\multicolumn{17}{r}{\small\itshape (continued)}\\
\toprule
\multirow{2}{*}{Model} & \multirow{2}{*}{Manipulation} & \multicolumn{3}{c}{Pre-self-report} & \multicolumn{3}{c}{Post-self-report} & \multicolumn{3}{c}{Performance (overall)} & \multicolumn{3}{c}{Effort} & \multicolumn{3}{c}{\# tokens} \\
\cmidrule(lr){3-5}\cmidrule(lr){6-8}\cmidrule(lr){9-11}\cmidrule(lr){12-14}\cmidrule(lr){15-17}
 &  & value & $T$ & $p$ & value & $T$ & $p$ & value & $T$ & $p$ & value & $T$ & $p$ & value & $T$ & $p$ \\
\midrule
\endhead
\multirow{11}{*}{GPT-4o} 
 & None & 81.574 & - & - & 90.318 & - & - & \textbf{5.765} & - & - & 5.825 & - & - & 511.715 & - & - \\
 & Competition & 80.832 & -1.014 & 0.345 & 91.994 & 6.268 & $p<0.001$ & 5.680 & -3.716 & $p<0.001$ & \underline{5.829} & 0.175 & 0.887 & 511.524 & -0.037 & 0.970 \\
 & Legacy & 81.613 & 0.139 & 0.898 & 88.499 & -6.324 & $p<0.001$ & 5.526 & -9.026 & $p<0.001$ & 5.477 & -11.629 & $p<0.001$ & 363.169 & -28.845 & $p<0.001$ \\
 & Purpose & 88.336 & 12.954 & $p<0.001$ & \textbf{96.171} & 19.520 & $p<0.001$ & 5.639 & -6.087 & $p<0.001$ & 5.805 & -0.903 & 0.403 & \underline{555.863} & 9.311 & $p<0.001$ \\
 & Encourage & 89.263 & 19.604 & $p<0.001$ & 95.568 & 16.102 & $p<0.001$ & \underline{5.688} & -3.553 & $p<0.001$ & \textbf{5.965} & 5.881 & $p<0.001$ & 532.204 & 4.368 & $p<0.001$ \\
 & Guilt & 86.905 & 17.199 & $p<0.001$ & 93.517 & 11.005 & $p<0.001$ & 5.676 & -3.984 & $p<0.001$ & 5.808 & -0.712 & 0.518 & 526.965 & 3.142 & 0.002 \\
 & Money & 92.484 & 28.383 & $p<0.001$ & 90.246 & -0.228 & 0.854 & 5.484 & -11.006 & $p<0.001$ & 5.584 & -8.993 & $p<0.001$ & 461.345 & -9.553 & $p<0.001$ \\
 & Money Loss & \underline{94.545} & 28.009 & $p<0.001$ & 94.526 & 14.910 & $p<0.001$ & 5.637 & -5.269 & $p<0.001$ & 5.829 & 0.142 & 0.898 & \textbf{598.181} & 14.635 & $p<0.001$ \\
 & Punish & \textbf{96.124} & 28.731 & $p<0.001$ & \underline{96.124} & 17.424 & $p<0.001$ & 5.315 & -14.977 & $p<0.001$ & 5.396 & -13.838 & $p<0.001$ & 391.348 & -20.051 & $p<0.001$ \\
 & Futility & 85.750 & 11.316 & $p<0.001$ & 93.576 & 12.097 & $p<0.001$ & 5.522 & -9.738 & $p<0.001$ & 5.753 & -2.841 & 0.005 & 482.178 & -6.127 & $p<0.001$ \\
 & Meaningless & 67.038 & -20.888 & $p<0.001$ & 87.109 & -5.858 & $p<0.001$ & 5.160 & -15.463 & $p<0.001$ & 5.145 & -16.427 & $p<0.001$ & 304.528 & -34.247 & $p<0.001$ \\
\midrule
\multirow{11}{*}{\makecell[l]{GPT-4o\\Mini}} 
 & None & 81.072 & - & - & 86.358 & - & - & \textbf{5.648} & - & - & \underline{5.699} & - & - & 504.292 & - & - \\
 & Competition & 82.403 & 2.900 & 0.005 & 82.403 & -7.546 & $p<0.001$ & 5.626 & -1.225 & 0.241 & 5.685 & -0.639 & 0.540 & 494.942 & -2.715 & 0.008 \\
 & Legacy & 81.993 & 5.285 & $p<0.001$ & 81.993 & -11.942 & $p<0.001$ & 5.549 & -4.727 & $p<0.001$ & 5.565 & -5.411 & $p<0.001$ & 414.011 & -22.029 & $p<0.001$ \\
 & Purpose & 85.007 & 8.890 & $p<0.001$ & 85.007 & -2.713 & 0.008 & 5.545 & -5.406 & $p<0.001$ & 5.605 & -4.192 & $p<0.001$ & 505.832 & 0.422 & 0.673 \\
 & Encourage & 85.399 & 20.334 & $p<0.001$ & 85.399 & -2.770 & 0.007 & 5.580 & -3.766 & $p<0.001$ & 5.653 & -2.136 & 0.037 & 492.413 & -3.863 & $p<0.001$ \\
 & Guilt & 85.193 & 21.452 & $p<0.001$ & 85.193 & -3.809 & $p<0.001$ & 5.616 & -1.817 & 0.077 & 5.694 & -0.267 & 0.789 & \underline{522.958} & 5.524 & $p<0.001$ \\
 & Money & \underline{88.853} & 22.546 & $p<0.001$ & \underline{88.853} & 8.039 & $p<0.001$ & 5.556 & -4.671 & $p<0.001$ & 5.622 & -3.185 & 0.002 & 500.851 & -0.871 & 0.409 \\
 & Money Loss & \textbf{89.643} & 26.373 & $p<0.001$ & \textbf{89.643} & 10.629 & $p<0.001$ & \underline{5.633} & -0.835 & 0.421 & \textbf{5.712} & 0.571 & 0.580 & \textbf{538.129} & 9.441 & $p<0.001$ \\
 & Punish & 81.652 & 0.851 & 0.416 & 81.652 & -6.509 & $p<0.001$ & 5.423 & -9.100 & $p<0.001$ & 5.458 & -8.826 & $p<0.001$ & 419.386 & -20.171 & $p<0.001$ \\
 & Futility & 78.088 & -14.887 & $p<0.001$ & 78.088 & -26.302 & $p<0.001$ & 5.382 & -11.288 & $p<0.001$ & 5.466 & -9.420 & $p<0.001$ & 431.042 & -18.325 & $p<0.001$ \\
 & Meaningless & 77.145 & -15.677 & $p<0.001$ & 77.145 & -21.755 & $p<0.001$ & 4.937 & -17.139 & $p<0.001$ & 4.906 & -17.403 & $p<0.001$ & 325.936 & -33.182 & $p<0.001$ \\
\midrule
\multirow{11}{*}{\makecell[l]{Llama\\3.1 8B}} 
 & None & 67.626 & - & - & 67.023 & - & - & \textbf{5.284} & - & - & \underline{5.487} & - & - & \underline{526.591} & - & - \\
 & Competition & 90.536 & 28.892 & $p<0.001$ & 74.901 & 18.246 & $p<0.001$ & \underline{5.276} & -0.248 & 0.821 & \textbf{5.518} & 0.990 & 0.336 & 512.192 & -2.474 & 0.015 \\
 & Legacy & 77.674 & 15.194 & $p<0.001$ & 52.976 & -17.710 & $p<0.001$ & 4.700 & -11.026 & $p<0.001$ & 4.787 & -12.015 & $p<0.001$ & 409.898 & -13.722 & $p<0.001$ \\
 & Purpose & \underline{91.090} & 32.256 & $p<0.001$ & \textbf{79.557} & 30.712 & $p<0.001$ & 5.245 & -1.342 & 0.192 & 5.487 & -0.021 & 0.983 & \textbf{527.485} & 0.153 & 0.888 \\
 & Encourage & 81.004 & 19.729 & $p<0.001$ & 74.030 & 14.449 & $p<0.001$ & 5.120 & -5.028 & $p<0.001$ & 5.381 & -3.330 & 0.001 & 434.539 & -15.454 & $p<0.001$ \\
 & Guilt & 77.878 & 17.284 & $p<0.001$ & \underline{75.446} & 14.471 & $p<0.001$ & 5.200 & -2.679 & 0.008 & 5.452 & -1.052 & 0.309 & 488.665 & -6.165 & $p<0.001$ \\
 & Money & 69.122 & 2.251 & 0.027 & 54.086 & -20.037 & $p<0.001$ & 5.004 & -7.521 & $p<0.001$ & 5.195 & -7.462 & $p<0.001$ & 455.421 & -10.201 & $p<0.001$ \\
 & Money Loss & \textbf{95.960} & 36.746 & $p<0.001$ & 62.510 & -4.501 & $p<0.001$ & 4.656 & -11.499 & $p<0.001$ & 4.894 & -9.907 & $p<0.001$ & 483.819 & -4.786 & $p<0.001$ \\
 & Punish & 85.296 & 22.395 & $p<0.001$ & 43.259 & -21.913 & $p<0.001$ & 3.750 & -21.860 & $p<0.001$ & 3.813 & -21.965 & $p<0.001$ & 303.058 & -21.406 & $p<0.001$ \\
 & Futility & 38.882 & -40.897 & $p<0.001$ & 50.463 & -27.841 & $p<0.001$ & 4.938 & -10.021 & $p<0.001$ & 5.240 & -7.287 & $p<0.001$ & 434.391 & -14.566 & $p<0.001$ \\
 & Meaningless & 37.024 & -36.888 & $p<0.001$ & 35.091 & -50.749 & $p<0.001$ & 4.953 & -8.911 & $p<0.001$ & 5.173 & -8.524 & $p<0.001$ & 446.920 & -13.123 & $p<0.001$ \\
\midrule
\multirow{11}{*}{\makecell[l]{Mistral-v0.3\\7B}} 
 & None & 80.954 & - & - & 86.409 & - & - & \underline{5.105} & - & - & 5.276 & - & - & 452.108 & - & - \\
 & Competition & 86.494 & 11.774 & $p<0.001$ & 85.777 & -1.237 & 0.270 & 5.037 & -2.424 & 0.023 & 5.250 & -0.973 & 0.394 & 441.461 & -1.989 & 0.065 \\
 & Legacy & 81.254 & 0.781 & 0.494 & 84.477 & -3.552 & $p<0.001$ & 5.092 & -0.471 & 0.679 & 5.261 & -0.557 & 0.635 & 422.527 & -5.724 & $p<0.001$ \\
 & Purpose & 86.014 & 11.219 & $p<0.001$ & \textbf{91.857} & 11.855 & $p<0.001$ & 5.028 & -2.912 & 0.006 & 5.235 & -1.626 & 0.141 & \underline{453.930} & 0.380 & 0.733 \\
 & Encourage & 88.088 & 13.273 & $p<0.001$ & \underline{89.385} & 8.240 & $p<0.001$ & 5.099 & -0.229 & 0.836 & 5.283 & 0.285 & 0.800 & 449.186 & -0.557 & 0.635 \\
 & Guilt & \textbf{89.778} & 14.995 & $p<0.001$ & 88.439 & 5.116 & $p<0.001$ & \textbf{5.108} & 0.110 & 0.913 & \underline{5.287} & 0.442 & 0.693 & 445.233 & -1.373 & 0.215 \\
 & Money & 84.996 & 9.176 & $p<0.001$ & 81.646 & -7.493 & $p<0.001$ & 5.010 & -3.394 & 0.001 & 5.198 & -2.848 & 0.007 & 439.911 & -2.373 & 0.025 \\
 & Money Loss & \underline{89.156} & 14.638 & $p<0.001$ & 87.311 & 1.457 & 0.186 & 5.063 & -1.505 & 0.172 & \textbf{5.288} & 0.472 & 0.679 & \textbf{470.168} & 3.134 & 0.003 \\
 & Punish & 86.436 & 12.449 & $p<0.001$ & 83.933 & -3.730 & $p<0.001$ & 4.987 & -4.164 & $p<0.001$ & 5.167 & -3.902 & $p<0.001$ & 426.461 & -4.807 & $p<0.001$ \\
 & Futility & 80.559 & -0.921 & 0.420 & 73.890 & -21.425 & $p<0.001$ & 4.835 & -9.111 & $p<0.001$ & 5.041 & -8.210 & $p<0.001$ & 387.821 & -12.375 & $p<0.001$ \\
 & Meaningless & 70.095 & -18.186 & $p<0.001$ & 74.622 & -16.665 & $p<0.001$ & 5.014 & -3.288 & 0.002 & 5.182 & -3.362 & 0.001 & 412.510 & -8.001 & $p<0.001$ \\
\midrule
\multirow{11}{*}{\makecell[l]{Gemini 2.0\\Flash}} 
 & None & 73.413 & - & - & 83.096 & - & - & 5.462 & - & - & 5.797 & - & - & 760.971 & - & - \\
 & Competition & 87.091 & 26.709 & $p<0.001$ & 87.546 & 14.924 & $p<0.001$ & \textbf{5.532} & 3.007 & 0.003 & \underline{5.896} & 3.961 & $p<0.001$ & 760.928 & -0.008 & 0.993 \\
 & Legacy & 80.299 & 16.424 & $p<0.001$ & 79.373 & -9.400 & $p<0.001$ & 5.272 & -5.588 & $p<0.001$ & 5.367 & -11.326 & $p<0.001$ & 520.288 & -27.551 & $p<0.001$ \\
 & Purpose & 82.528 & 18.530 & $p<0.001$ & 88.721 & 17.495 & $p<0.001$ & 5.453 & -0.401 & 0.717 & 5.827 & 1.344 & 0.201 & 763.356 & 0.523 & 0.632 \\
 & Encourage & 84.132 & 25.489 & $p<0.001$ & 87.107 & 13.920 & $p<0.001$ & \underline{5.467} & 0.237 & 0.830 & 5.883 & 3.534 & $p<0.001$ & 755.909 & -1.079 & 0.312 \\
 & Guilt & 79.545 & 15.854 & $p<0.001$ & 89.745 & 22.373 & $p<0.001$ & 5.278 & -6.500 & $p<0.001$ & 5.717 & -2.731 & 0.008 & 696.345 & -8.942 & $p<0.001$ \\
 & Money & 84.378 & 23.955 & $p<0.001$ & 88.379 & 13.761 & $p<0.001$ & 5.372 & -3.837 & $p<0.001$ & 5.890 & 3.872 & $p<0.001$ & \underline{829.812} & 12.305 & $p<0.001$ \\
 & Money Loss & \underline{94.010} & 36.422 & $p<0.001$ & \underline{91.487} & 24.644 & $p<0.001$ & 5.351 & -4.383 & $p<0.001$ & \textbf{5.924} & 4.994 & $p<0.001$ & \textbf{845.327} & 15.185 & $p<0.001$ \\
 & Punish & \textbf{94.924} & 34.449 & $p<0.001$ & \textbf{94.924} & 34.194 & $p<0.001$ & 5.247 & -6.656 & $p<0.001$ & 5.461 & -9.400 & $p<0.001$ & 583.299 & -21.321 & $p<0.001$ \\
 & Futility & 64.434 & -19.895 & $p<0.001$ & 60.446 & -45.126 & $p<0.001$ & 5.283 & -6.757 & $p<0.001$ & 5.621 & -6.163 & $p<0.001$ & 638.216 & -17.661 & $p<0.001$ \\
 & Meaningless & 31.347 & -49.709 & $p<0.001$ & 8.433 & -102.024 & $p<0.001$ & 5.116 & -8.056 & $p<0.001$ & 5.220 & -12.385 & $p<0.001$ & 460.888 & -31.912 & $p<0.001$ \\
\midrule
\multirow{11}{*}{All} 
 & None & 75.235 & - & - & 81.292 & - & - & \textbf{5.407} & - & - & 5.564 & - & - & 541.289 & - & - \\
 & Competition & 84.156 & 28.856 & $p<0.001$ & 83.224 & 10.131 & $p<0.001$ & \underline{5.388} & -1.607 & 0.118 & \textbf{5.582} & 1.437 & 0.159 & 533.838 & -3.397 & $p<0.001$ \\
 & Legacy & 79.229 & 22.042 & $p<0.001$ & 75.615 & -24.699 & $p<0.001$ & 5.153 & -15.167 & $p<0.001$ & 5.199 & -19.648 & $p<0.001$ & 411.733 & -41.086 & $p<0.001$ \\
 & Purpose & 85.145 & 34.965 & $p<0.001$ & \textbf{86.778} & 29.213 & $p<0.001$ & 5.339 & -6.125 & $p<0.001$ & 5.544 & -1.718 & 0.095 & \underline{549.289} & 3.897 & $p<0.001$ \\
 & Encourage & 83.938 & 36.437 & $p<0.001$ & 85.025 & 25.220 & $p<0.001$ & 5.336 & -6.388 & $p<0.001$ & \underline{5.578} & 1.171 & 0.252 & 521.467 & -9.193 & $p<0.001$ \\
 & Guilt & 82.292 & 33.187 & $p<0.001$ & \underline{85.111} & 23.632 & $p<0.001$ & 5.328 & -6.706 & $p<0.001$ & 5.536 & -2.253 & 0.027 & 523.617 & -7.121 & $p<0.001$ \\
 & Money & 82.446 & 28.532 & $p<0.001$ & 79.208 & -9.429 & $p<0.001$ & 5.220 & -14.092 & $p<0.001$ & 5.421 & -10.189 & $p<0.001$ & 523.285 & -7.151 & $p<0.001$ \\
 & Money Loss & \textbf{91.287} & 46.163 & $p<0.001$ & 84.215 & 12.284 & $p<0.001$ & 5.184 & -15.345 & $p<0.001$ & 5.427 & -8.949 & $p<0.001$ & \textbf{568.615} & 10.490 & $p<0.001$ \\
 & Punish & \underline{87.446} & 41.750 & $p<0.001$ & 79.022 & -6.884 & $p<0.001$ & 4.866 & -27.343 & $p<0.001$ & 4.963 & -28.771 & $p<0.001$ & 410.775 & -37.639 & $p<0.001$ \\
 & Futility & 68.118 & -30.346 & $p<0.001$ & 70.155 & -54.662 & $p<0.001$ & 5.156 & -17.582 & $p<0.001$ & 5.376 & -13.585 & $p<0.001$ & 464.966 & -26.749 & $p<0.001$ \\
 & Meaningless & 55.518 & -59.762 & $p<0.001$ & 55.927 & -97.687 & $p<0.001$ & 4.995 & -19.674 & $p<0.001$ & 5.075 & -22.027 & $p<0.001$ & 380.896 & -45.091 & $p<0.001$ \\
\midrule
\end{longtable}
}
\end{landscape}
\restoregeometry
\endgroup
\end{appendices}

\end{document}